\documentclass[journal,onecolumn,12pt, draftclsnofoot, peerreview]{IEEEtran}

\usepackage{amsmath,amsfonts,amssymb}
\usepackage{algorithmic}
\usepackage{array}
\usepackage[caption=false]{subfig} %
\usepackage{textcomp}
\usepackage{stfloats}
\usepackage{url}
\usepackage{verbatim}
\usepackage{graphicx}
\def\BibTeX{{\rm B\kern-.05em{\sc i\kern-.025em b}\kern-.08em
    T\kern-.1667em\lower.7ex\hbox{E}\kern-.125emX}}
\usepackage{balance}

\usepackage{pifont}
\newcommand{\crossmark}{\ding{55}}  %

\IEEEoverridecommandlockouts 

\usepackage{tabularx}
\usepackage{multirow}

\usepackage{booktabs}
\usepackage{xspace}
\DeclareMathOperator*{\argmax}{arg\,max}

\usepackage{cite}

\usepackage{lineno} %

\makeatletter
\DeclareRobustCommand\onedot{\futurelet\@let@token\@onedot}
\def\@onedot{\ifx\@let@token.\else.\null\fi\xspace}

\def\eg{\emph{e.g}\onedot} 
\def\ie{\emph{i.e}\onedot}

\def\etal{\emph{et al}\onedot}
\makeatother

\usepackage{eso-pic}
\usepackage{url}
\AddToShipoutPicture*{%
     \AtTextUpperLeft{%
         \put(-3.5,10){
           \begin{minipage}{\textwidth}
              \scriptsize
              \MakeUppercase{Preprint version of an IEEE Journal of Oceanic Engineering article; \newline final version available at} \url{https://doi.org/10.1109/JOE.2025.3625691} 
              \newline
           \end{minipage}}%
     }%
}

\begin{document}

\title{Human-in-the-Loop Segmentation of Multi-species Coral Imagery}

\author{Scarlett Raine$^{1}$ \IEEEmembership{Member, IEEE}, Ross Marchant$^{2}$, \\ Brano Kusy$^{3}$ \IEEEmembership{Member, IEEE}, Frederic Maire$^{1}$, Niko S\"{u}nderhauf$^{1}$ \IEEEmembership{Member, IEEE} \\ and Tobias Fischer$^{1}$ \IEEEmembership{Senior Member, IEEE}
\thanks{$^{1}$QUT Centre for Robotics, Australia \tt \emph{\{sg.raine, f.maire, niko.suenderhauf, tobias.fischer\}@qut.edu.au} \\
\normalfont $^{2}$Image Analytics, Australia \tt \emph{ross.g.marchant@gmail.com} \\
\normalfont $^{3}$CSIRO Data61, Australia \tt \emph{brano.kusy@csiro.au}
}}

\markboth{IEEE Journal of Oceanic Engineering}
{S. Raine, R. Marchant, \MakeLowercase{\textit{(et al.)}: Human-in-the-Loop Segmentation of Multi-species Coral Imagery}}

\maketitle

\begin{abstract}
    Marine surveys by robotic underwater and surface vehicles result in substantial quantities of coral reef imagery, however labeling these images is expensive and time-consuming for domain experts. Point label propagation is a technique that uses existing images labeled with sparse points to create augmented ground truth data, which can be used to train a semantic segmentation model. In this work, we show that recent advances in large foundation models facilitate the creation of augmented ground truth masks using only features extracted by the denoised version of the DINOv2 foundation model and K-Nearest Neighbors (KNN), without any pre-training. For images with extremely sparse labels, we use human-in-the-loop principles to enhance annotation efficiency: if there are 5 point labels per image, our method outperforms the prior state-of-the-art by 19.7\% for mIoU. When human-in-the-loop labeling is not available, using the denoised DINOv2 features with a KNN still improves on the prior state-of-the-art by 5.8\% for mIoU \mbox{(5 grid points)}. On the semantic segmentation task, we outperform the prior state-of-the-art by 13.5\% for mIoU when only 5 point labels are used for point label propagation.  Additionally, we perform a comprehensive study into the number and placement of point labels, and make several recommendations for improving the efficiency of labeling images with points.
\end{abstract}

\newpage

\section{Introduction}
\label{sec:intro}  

\begin{figure}[t]
\includegraphics[width=\columnwidth, clip, trim=2.8cm 1.4cm 2.8cm 2.0cm]{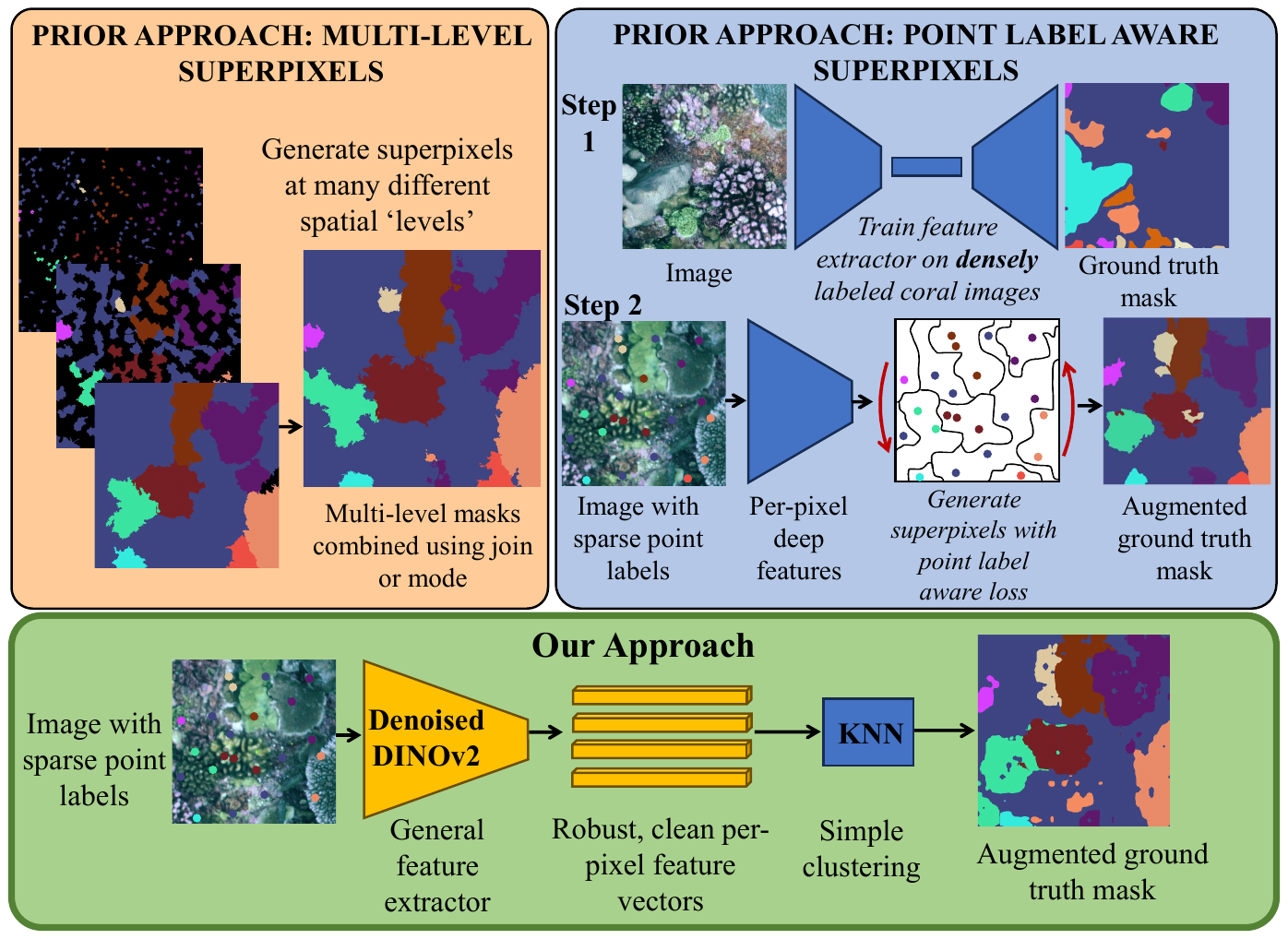}
\caption{The proposed point label propagation technique utilizes the DINOv2 foundation model without any fine-tuning to create augmented ground truth masks for intricate coral images. \textbf{Top left:} Previous methods depended on layering superpixels that contained point labels \cite{alonso2018semantic, alonso2019coralseg, pierce2020reducing}. \textbf{Top right:} A more recent approach involved pre-training a CNN feature extractor on densely labeled coral images and propagating point labels with a custom, point label aware superpixel function \cite{raine2022point}. \textbf{Bottom:} In contrast, our approach employs KNN to group features derived from the denoised version of the DINOv2 foundation model without further training on coral imagery.
}
\label{fig:frontpage}
\end{figure}
  
\IEEEPARstart{I}{nformed} marine ecosystem management relies on data gathered at various spatial and temporal scales \cite{ditria2022artificial}. Marine surveys are increasingly being conducted using autonomous underwater and surface vehicles \cite{dunbabin2020uncrewed, gregorek2023long, mizuno2019development}.  However, these methods produce vast amounts of seafloor images that need to be analyzed to produce meaningful statistics such as coverage estimates of various substrates and coral species \cite{mahmood2018deep, xu2019deep, runyan2022automated}. 

One method for automated image analysis is semantic segmentation, a computer vision task that predicts the class of every pixel in a query image~\cite{guo2018review}. In recent years, it has become common practice to train deep learning models for performing semantic segmentation.  These models are often trained using full supervision, \ie on pairs of images and dense ground truth masks, where the mask is comprised of a class label for every corresponding pixel in the paired image~\cite{zhang2019survey}. These dense masks are typically created by human experts who manually label each pixel.  

However, underwater imagery has visually unique characteristics which complicate the process of labeling images.  Coral images frequently exhibit high complexity, and feature unclear boundaries, significant color and texture variation among species, and low clarity \cite{fu2023masnet, li2024survey, sun2018transferring, jin2017deep}. The complex and intricate nature of coral images and the challenges in correctly classifying coral species necessitate annotation of images by marine scientists, thus hindering use of standard computer vision annotation tools like the crowd-sourcing provider Amazon Turk\footnote{\url{https://www.mturk.com/}}.

Historically, marine scientists have labeled underwater images using the Coral Point Count (CPC) \cite{kohler2006coral} methodology, in which a scientist specifies the class of a set quantity of randomly distributed or grid-spaced sparse pixels in each image. We call these annotated pixels \textit{point labels} \cite{raine2022point}. The CPC method typically involves labeling 100-300 point labels per image.  The extremely sparse label setting that we target in this paper is critical for ecologists who often have limited budget and time constraints for monitoring projects and are unable to label 100-300 points per image~\cite{gonzalez2020monitoring}. The sparse label setting can also occur during field expeditions, when ecologists might be performing surveys in a new location, for a specific task or to detect certain species.  In this case, it is necessary to quickly train, re-train or fine-tune models to perform semantic segmentation, often overnight or between transect surveys, but it is prohibitively time-consuming and costly for domain experts to label every pixel or even 300 pixels in each image~\cite{gonzalez2020monitoring}.  By contrast, it could be feasible for experts to quickly label 5 points per image and use a point label propagation algorithm like ours to create dense augmented ground truth for training or fine-tuning a semantic segmentation model.

Although there are large quantities of CPC data available in point grid and random formats \cite{beijbom2012automated, gonzalez2014catlin}, the most efficient way of labeling points for training deep learning models to perform semantic segmentation has not been investigated, and could have significant impacts for annotation costs.

Recently, superpixel algorithms based on RGB color values \cite{alonso2018semantic, alonso2019coralseg, pierce2020reducing} and deep features \cite{raine2022point} have been leveraged for propagating point labels into dense, pixel-wise augmented ground truth masks.  These masks are then used as the supervisory signal when training deep neural networks to perform semantic segmentation of underwater images. Raine \etal \cite{raine2022point} presented an innovative point label aware approach to superpixels, clustering pixels into segments using the per-pixel deep CNN features and the RGB values. Although this algorithm advanced the state-of-the-art, the deep features are extracted by a CNN trained on coral imagery, and experienced performance issues when only a small number of point labels were available.

In this study, we address the scenario where only a very small number of labels are available. This situation is crucial because marine survey projects frequently have constrained budgets for data labeling \cite{ditria2022artificial}. Moreover, processing survey data often requires swiftly fine-tuning and retraining models in the field to adapt to new locations, species or environmental conditions \cite{ditria2022artificial}. In this work, we propose using the DINOv2 foundation model \cite{oquab2024dinov, yang2024denoising} to generate per-pixel embeddings. We then employ the straightforward K-Nearest Neighbor algorithm to create the augmented ground truth, surpassing the state-of-the-art with a limited number of point labels. Additionally, we show further performance enhancements by implementing a human-in-the-loop point selection strategy, utilizing the expertise of human experts to reduce uncertainty in the embedding space of the KNN.

This work establishes new methodological approaches for foundation model adaptation in specialized domains, demonstrating how general representations can be effectively combined with domain-specific expert knowledge for efficient annotation.  For the first time, we address the problem of label efficiency in the extremely sparse label setting (Fig.~\ref{fig:frontpage}) through principled human-machine collaboration. Our contributions are summarized as follows:
\begin{enumerate}
    \item \textbf{Foundation Model Adaptation:} We propose using a general-purpose foundation model to produce per-pixel deep features for coral images, demonstrating that these features are discriminative without training or fine-tuning on coral imagery. When combined with the basic \mbox{K-Nearest} Neighbors algorithm, these features are sufficient for creating accurate augmented ground truth masks, eliminating the need for custom superpixel algorithms.
    \item \textbf{Human-in-the-Loop Framework:} We present a novel human-in-the-loop labeling regime for  the extremely sparse point label setting, \ie 5-25 points per image, which integrates the expertise of the marine scientist with the model's introspective uncertainty to identify informative locations for point labels. Our approach outperforms the prior state-of-the-art method by 14.2\% pixel accuracy and 19.7\% mIoU for the 5 point label setting, and by 8.9\% and 18.3\% when there are 10 points labeled in each image.
    \item \textbf{Robust Performance without Human Guidance:} If human-in-the-loop labeling is not available, we find that leveraging the DINOv2 denoised features~\cite{oquab2024dinov, yang2024denoising} with a KNN results in improvements over the stage-of-the-art for propagation of small quantities of point labels per image. We see pixel accuracy improve by 2.7\% and an improvement of 5.8\% for mIoU (5 point labels per image); and 2.3\% in pixel accuracy and 10.0\% in mIoU (10 point labels per image) on  UCSD Mosaics~\cite{edwards2017large, alonso2019coralseg}.
    \item \textbf{End-to-End Semantic Segmentation:} These improvements in point label propagation are reflected in the semantic segmentation task. A DeepLabv3+ model~\cite{chen2017rethinking} trained on augmented ground truth masks generated using DINOv2~\cite{oquab2024dinov, yang2024denoising}, KNN and our human-in-the-loop labeling regime significantly outperforms the prior works: pixel accuracy improves by 8.8\% and mIoU improves by 13.5\% (5 point labels per UCSD Mosaics~\cite{edwards2017large, alonso2019coralseg} image).
    \item \textbf{Comprehensive Analysis:} We conduct comprehensive experiments to assess the impact of the quantity and placement of point labels on the point propagation task, offering valuable recommendations for efficient annotation.
\end{enumerate}

\begin{figure*}[t]
\includegraphics[width=\textwidth, clip, trim=0cm 8cm 0cm 0cm]{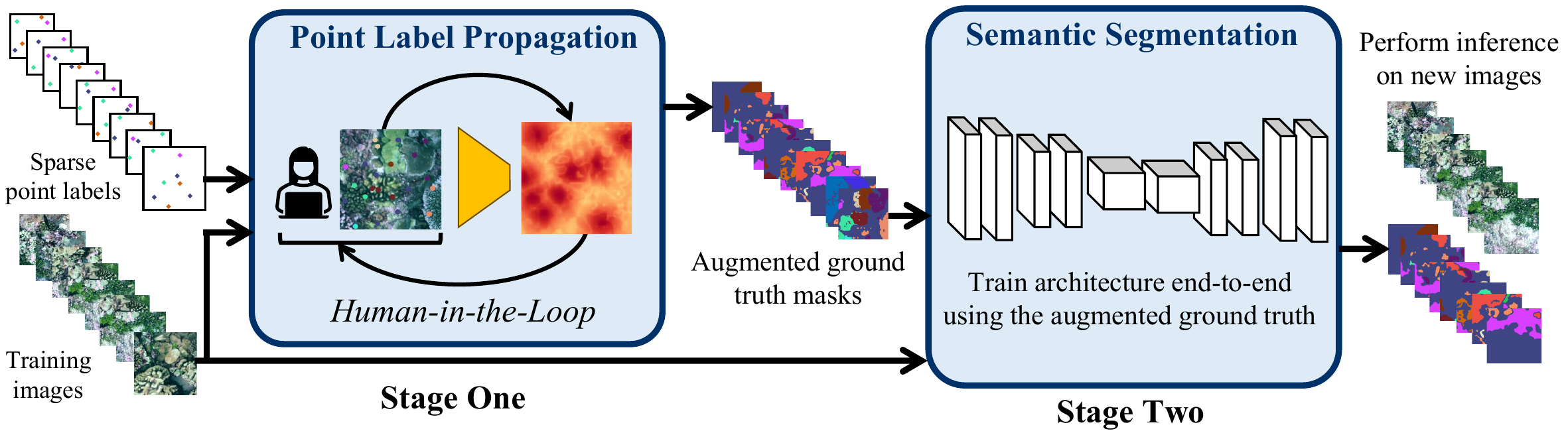}
\caption{Schematic of Full Semantic Segmentation Framework.  We combine the human-in-the-loop point label propagation approach (Stage One) with a semantic segmentation architecture (Stage Two).  The semantic segmentation architecture is trained on the augmented ground truth masks to enable inference on unlabeled imagery. Refer to Section~\ref{subsec:ablations} for an analysis of various semantic segmentation architectures.}
\label{fig:segpipeline}
\end{figure*}

An earlier version of this work was presented as a conference paper~\cite{raine2024human}. The major contributions compared to the conference version are:
\begin{enumerate}
    \item Extension of the proposed algorithm to a complete two-stage semantic segmentation pipeline and thorough analysis of the second stage:
    \begin{itemize}
        \item Addition of Stage Two enables inference on unlabeled images at test time
        \item Implementation of a semantic segmentation model for automated analysis of coral imagery (described in Section~\ref{subsec:segmentation} and Section~\ref{subsubsec:Implementation-seg})
        \item Integration of both stages into a cohesive framework for practical application (shown in Fig.~\ref{fig:segpipeline})
    \end{itemize}
    
    \item Comprehensive experimental evaluation and analysis:        
    \begin{itemize}            
        \item Doubled the quantitative and qualitative results presented (Tables~\ref{tab:stage2} and~\ref{tab:seg}, and expansion of Figs.~\ref{fig:qualitativeresults} and~\ref{fig:denoise})           
        \item Evaluation of various semantic segmentation architectures and their backbones (Section~\ref{subsubsec:seg} and Section~\ref{subsec:ablation-seg})           
        \item Comparative analysis of inference speeds for practical deployment considerations (Table~\ref{tab:seg})
    \end{itemize}    
    
    \item Expanded ablation studies for the point label propagation approach:        
    \begin{itemize}            
        \item New analysis of clustering parameters ($k$ values) across different quantities of point labels (Fig.~\ref{fig:k-ablation})
        \item Investigation of minimum number of human-labeled initial points required (Fig.~\ref{fig:human-points-ablation})         
        \item Extended evaluation of feature extractors, including DINOv2 variants with Registers~\cite{oquab2024dinov, darcet2023vision} (Fig.~\ref{fig:denoise})
    \end{itemize}    
    
    \item Enhanced visual results and in-depth discussion:      
    \begin{itemize}            
        \item New demonstrations of foundation model limitations on complex coral imagery (Figs.~\ref{fig:sam} and~\ref{fig:dinov2} provide analysis of Segment Anything 2~\cite{kirillov2023segment, ravi2024sam} and DINOv2~\cite{oquab2024dinov} results when used out-of-the-box)              
        \item Analysis of positional embedding artifacts in the feature space (Fig.~\ref{fig:ViT-noise})
        \item Additional examples illustrating dataset cleaning (Fig.~\ref{fig:ucsd}) and visualization of ground truth masks (Fig.~\ref{fig:legend})     \item Additional discussion of impact, limitations and future work for the proposed method (Section~\ref{sec:discussion}). 
    \end{itemize}

\end{enumerate}

\noindent To foster future research in this area, we make our code publicly available: \url{https://github.com/sgraine/HIL-coral-segmentation}.

\section{Related Work}
\label{sec:related}
Performing marine surveys with autonomous surface and underwater vehicles enables the collection of significant quantities of images \cite{li2022real, mou2022reconfigurable, dunbabin2020uncrewed, gregorek2023long}. Automating the analysis of underwater imagery requires innovative approaches based on deep learning, robotics, computer vision and specialized expertise in underwater ecosystems \cite{xu2019deep, gonzalez2016scaling}. This section reviews and analyzes prior methods for semantic segmentation of underwater imagery and point label propagation (Section~\ref{subsec:underwater_seg}), advances in large foundation models (Section~\ref{subsec:foundation_models}), and human-in-the-loop fundamentals (Section~\ref{subsec:human_in_the_loop}).

\subsection{Semantic Segmentation of Coral Imagery}
\label{subsec:underwater_seg}

Underwater semantic segmentation is challenging due to the dynamic, unrestricted environment of the ocean~\cite{jin2017deep}.  Typically images are deteriorated by noise, turbidity, scattering and attenuation of sunlight, low-light conditions, blur, and changes in coloration due to depth~\cite{sun2018transferring, jin2017deep}.  For example, image characteristics vary significantly both within and between datasets due to lighting (whether artificial lighting or natural lighting is employed), camera settings used, and post-processing of images.  Coral species often exhibit strong visual similarity between different species, and are typically intricate and highly textured \cite{li2024survey,beijbom2012automated,gonzalez2020monitoring}. Underwater image characteristics, along with the lack of distinct ``objectness" for overlapping corals, make the underwater semantic segmentation task a uniquely challenging problem.

Many works which perform fully supervised segmentation of underwater coral imagery train the model with images and their corresponding densely labeled, pixel-wise ground truth masks \cite{ziqiang2023coralvos, zhang2024cnet, zhang2022deep, sui2022automated, zhong2023combining, furtado2023deolhonoscorais, islam2020semantic, sui2022automated}. Some approaches have proposed interactive labeling tools which use deep learning to assist annotation, but these approaches do not target label efficiency or propose informative label locations \cite{sui2022automated, langenkamper2017biigle,beijbom2015towards}. The Machine learning Assisted Image Annotation (MAIA) method proposes novel regions for labeling, but relies on an autoencoder pre-trained on coral images \cite{zurowietz2018maia}. TagLab is a data labeling tool designed to speed up the pixel-wise annotation of large orthoimages\footnote{Orthoimages or orthomosaics are geometrically corrected images which have been adjusted to remove distortions due to the camera, water and seafloor topography. Orthoimages can be generated from a series of overlapping images taken by an underwater camera or robotic vehicle and then joined together into a mosaic to cover a larger area \cite{urbina2021method}.}, however it is based on a model trained using 15,000 coral images with dense ground truth labels \cite{pavoni2021taglab}. Recently, a human-in-the-loop iterative labeling approach was proposed which combines expert and crowd-sourced non-expert annotations, however their method is only for bounding box detection of marine species \cite{zhang2023iterative}.

There are limited algorithms in the literature for weakly supervised segmentation of corals \cite{yu2019weakly, yu2019fast, alonso2018semantic, alonso2019coralseg, pierce2020reducing, raine2022point}.  These rely on custom-designed superpixel algorithms which propagate sparse point labels to obtain corresponding pixel-wise ground truth.  The multi-level superpixel method generates superpixel segments informed by the RGB features. This algorithm repeatedly generates the segments at a variety of spatial scales, each time labeling each segment as the class of the point label inside before combining the various ``levels" of superpixels into a single augmented ground truth \cite{alonso2018semantic, alonso2019coralseg, pierce2020reducing}.  More recently, Point Label Aware Superpixels \cite{raine2022point} was proposed and introduced a novel superpixel algorithm which clusters pixels with deep features and creates superpixel segments informed directly by the point labels. 

\begin{figure}[t]
\centering
\centerline{\includegraphics[width=170mm, clip, trim=0.8cm 5cm 1cm 3.25cm]{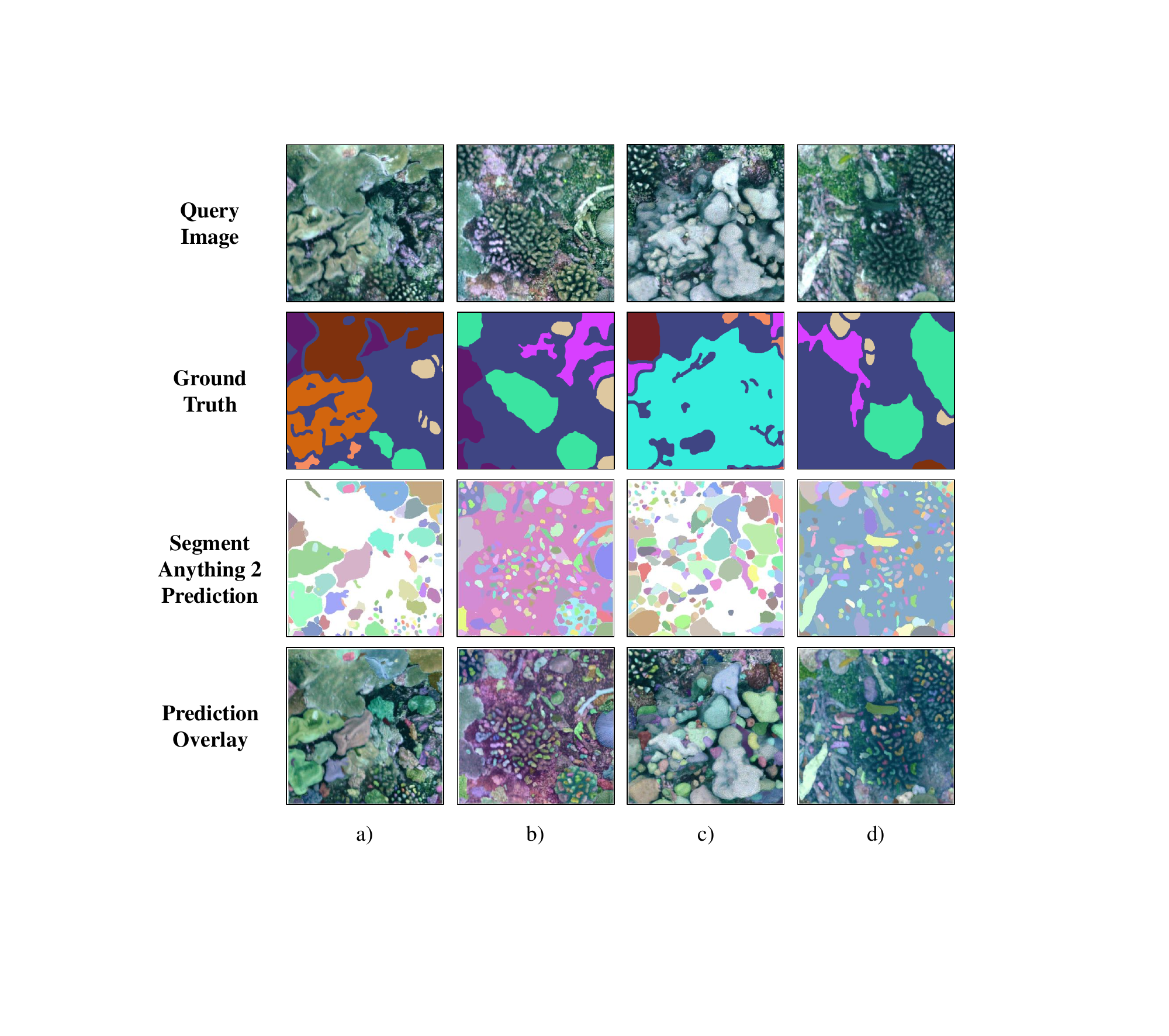}} 
\caption[Segment Anything 2 Performance]{Instance segmentation by Segment Anything 2~\cite{ravi2024sam} on UCSD Mosaics~\cite{edwards2017large, alonso2019coralseg} coral images. This figure shows that Segment Anything 2 does not produce usable segments in these images, and also does not provide any class predictions for the segments. Refer to Fig.~\ref{fig:legend} for the color legend for the ground truth masks.}
\label{fig:sam}
\end{figure}

There have been recent advances in large, general instance segmentation models \eg the Segment Anything (SAM)~\cite{kirillov2023segment} and Segment Anything 2~\cite{ravi2024sam} models.  It is important to highlight the difference between the instance segmentation and semantic segmentation tasks.  Instance segmentation groups pixels in an image into distinct segments without classifying the object.  Semantic segmentation instead predicts the presence of specific classes of interest, and localizes those objects by denoting the pixels which belong to each class.  We evaluate the performance of SAM 2 on coral imagery in Fig.~\ref{fig:sam}, which demonstrates that the output of these models does not provide class information, and that the segments produced do not align well with the intended ground truth masks.

Previous superpixel methods depend on the availability of a sufficient number of points and suffer from degraded performance in very sparse settings. There is potential to utilize the recent innovations in large foundation models for a more general, simplified approach to point label propagation. 

\subsection{Large Foundation Models}
\label{subsec:foundation_models}
Recently there has been considerable research effort towards developing large foundation models which learn robust, discriminative feature embeddings that are task-agnostic \cite{oquab2024dinov, kirillov2023segment, zou2024segment}. Foundation models are trained on extensive datasets and designed to acquire highly generalized representations, enabling them to apply their knowledge to tasks and data beyond the training scope \cite{kirillov2023segment}.

One such foundation model, DINOv2~\cite{oquab2024dinov}, was trained on 142 million images, and is based on a Vision Transformer (ViT) model~\cite{dosovitskiy2020image} trained with a discriminative self-supervised loss function~\cite{oquab2024dinov}.  This loss function combines an image-level objective function for features extracted from a student-teacher framework and a patch-level objective, in which patches from the input image are masked and the model must predict the missing image regions~\cite{oquab2024dinov}.  There have been extensions and improvements of the DINOv2 architecture: the Denoising Vision Transformers approach~\cite{yang2024denoising} removes positional artefacts from deep features, and can be applied on top of any ViT; and Darcet \etal~\cite{darcet2023vision} propose using registers, which are additional tokens the model can use to store global information, therefore reducing artefacts in ViT features.

Some studies focus on tailoring foundation models to specific tasks or contexts, such as for recognition of a user’s pet \cite{zhang2023personalize}, or by training a decoder or adapter on the desired task \cite{chen2023adapting}. In the context of plant phenotyping, modified foundation models were evaluated for instance segmentation, leaf counting and disease classification, however the methods designed specifically for these tasks outperformed the modified foundation models \cite{chen2023adapting}. In medical image analysis, \cite{baharoon2023towards} show that DINOv2 has cross-task generalizability and report competitive results when its features are applied with KNN for disease classification. Other research has explored self-supervised object localization without using labels \cite{simeoni2021localizing}, but these methods do not achieve segmentation of the entire image.

\begin{figure}[t]
\centering
\centerline{\includegraphics[width=170mm, clip, trim=0.8cm 2cm 1cm 3.25cm]{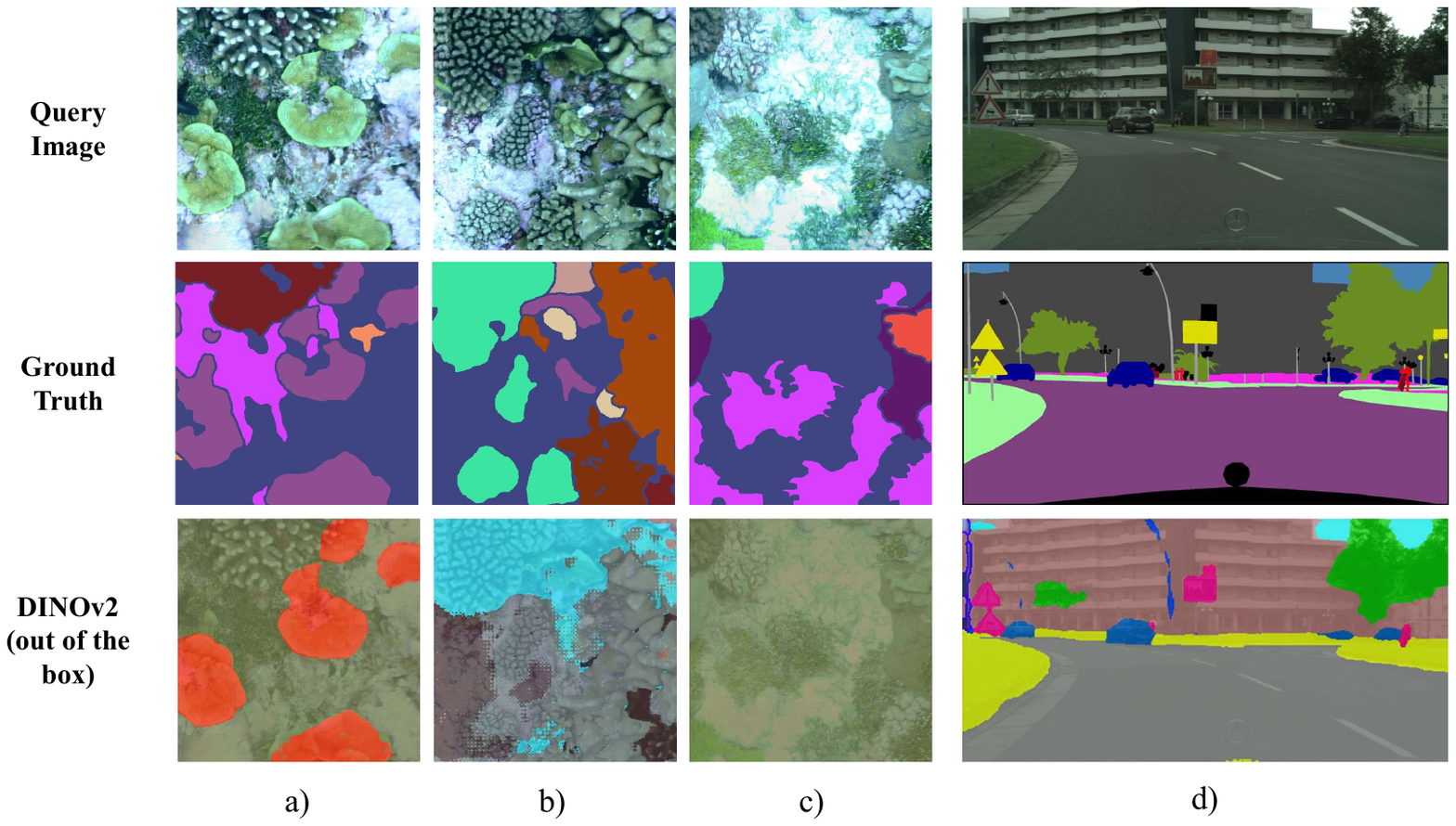}} 
\caption[DINOv2 Performance]{A comparison of segmentation by DINOv2~\cite{oquab2024dinov} on UCSD Mosaics~\cite{edwards2017large, alonso2019coralseg} coral images (a, b, and c) and an image from the Cityscapes~\cite{cordts2016cityscapes} dataset (d).  Note that in example c), the DINOv2 model does not produce any segments.  This demonstrates that DINOv2 cannot be directly used in domain-specific applications. This work instead proposes leveraging the deep feature space in combination with a nearest neighbor classifier to perform point label propagation.}
\label{fig:dinov2}
\end{figure}

While some works have explored the utility of foundation models for specialized problems \cite{huix2024natural, baharoon2023towards, chen2023adapting}, using DINOv2 for the task of underwater coral segmentation has not been investigated.  As highlighted by Section~\ref{subsec:underwater_seg}, coral imagery has unique and challenging visual characteristics, including detailed textures, poorly defined boundaries, and overlapping species, as seen in Fig.~\ref{fig:dinov2} \cite{raine2022point, beijbom2012automated}.  When applied to common imagery, such as on an example image from the Cityscapes~\cite{cordts2016cityscapes} dataset, the DINOv2 model is able to effectively segment image, however when used directly on the coral images, the model is not able to generate useful segments (Fig.~\ref{fig:dinov2}).  This suggests that out-of-the-box usage of DINOv2 on this domain-specific task is not effective. However, our work investigates whether the feature representation of the foundation model DINOv2, trained on general images, can generate meaningful feature embeddings for coral images. We propose leveraging the deep embedding space of DINOv2 alongside sparse domain expert point labels for point label propagation in coral imagery. 

\subsection{Human-in-the-Loop}
\label{subsec:human_in_the_loop}

Human-in-the-Loop refers to machine learning algorithms and pipelines that enable humans to directly provide feedback to and engage with a model \cite{mosqueira2023human}. More specifically, the Human-in-the-Loop subfield of Interactive Machine Learning outlines a framework where control is shared between humans and models; humans provide input to the system in a direct, frequent and interactive manner \cite{jiang2019recent, mosqueira2023human}.

Although models have previously been used to predict labels on unseen data in the field of ecology \cite{chen2021new, kellenberger2020aide}, the application of foundation models within an interactive labeling framework for coral point label propagation has not yet been explored.

To our knowledge, there is no existing approach in the literature for semantic segmentation of multi-species coral imagery that integrates the power of task-agnostic foundation models with domain-specific, annotation-efficient labeling. This presents an opportunity to reduce the time and costs associated with manual annotation of complex, domain-specific imagery, while improving point label propagation when there are extremely limited labels available.

\section{Method}
\label{sec:method}

\subsection{Method Overview}
\label{subsec:methodoverview}

This section provides an overview of our proposed point label propagation approach. We leverage the denoised version of the DINOv2 foundation model by \cite{yang2024denoising}, based on \cite{oquab2024dinov}.  We cluster pixels in the deep embedding space with K-Nearest Neighbors, yielding the augmented ground truth mask.  

Our method expects a photo-quadrat coral image and a collection of sparse point labels as input, producing a dense augmented ground truth mask as output. The input point labels may be randomly distributed across the image, arranged evenly in a grid\footnote{If the grid cannot evenly accommodate the total number of points, the nearest possible arrangement is used; for example, in the case of 5 point labels, the point labels are spaced in a 2x2 grid with a single point in the center. For the 10 point label case, we use 3 rows of points, where the first and third rows contain 3 point labels and the middle row has 4 evenly spaced points.}, or designated by our proposed human-in-the-loop framework (Section~\ref{subsec:hil}).

From our image, we generate a set of feature vectors, each 768 in length, extracted from the denoised DINOv2 feature extractor \cite{yang2024denoising}. This extractor produces a deep feature for every 14x14 pixel patch in the input image. Additionally, it generates a `CLS' token for the entire image, which is not used in our method. We then perform spatial upsampling on the feature vectors using bilinear interpolation to create a deep feature vector for each pixel in the input image, after which we L2 normalize the per-pixel feature vectors.

We obtain a set of sparse labeled pixels $L$ and retain the normalized feature embeddings \{$\textbf{v}_1$, \dots, $\textbf{v}_l$, \dots, $\textbf{v}_L$\} for these pixels. In addition, we store $X=\{(x_1,y_1),\dots,(x_L,y_L)\}$ where $(x_l,y_l)$ are the pixel coordinates of $\textbf{v}_l$. We find the cosine similarity of the feature embedding $\textbf{v}_l$ for $l \in \{1,\dots,L\}$ and the feature embedding $\textbf{v}_{p}$ for every other pixel $p \neq l$ in the image:
\begin{equation}
    \label{eq:cosine}
        \mathrm{sim(}\textbf{v}_{p}, {\textbf{v}}_l) = \textbf{v}_{p} \cdot {\textbf{v}}_l.
\end{equation} 

We obtain a dense ground truth mask by performing K-Nearest Neighbors with $k=1$.  Note that although we evaluated various values for $k$  this did not lead to any improvement for $k>1$, as discussed in Section~\ref{subsec:ablations} and in Fig.~\ref{fig:k-ablation}. 

\begin{figure*}[t]
\includegraphics[width=\textwidth, clip, trim=0.2cm 3.5cm 0.2cm 3.5cm]{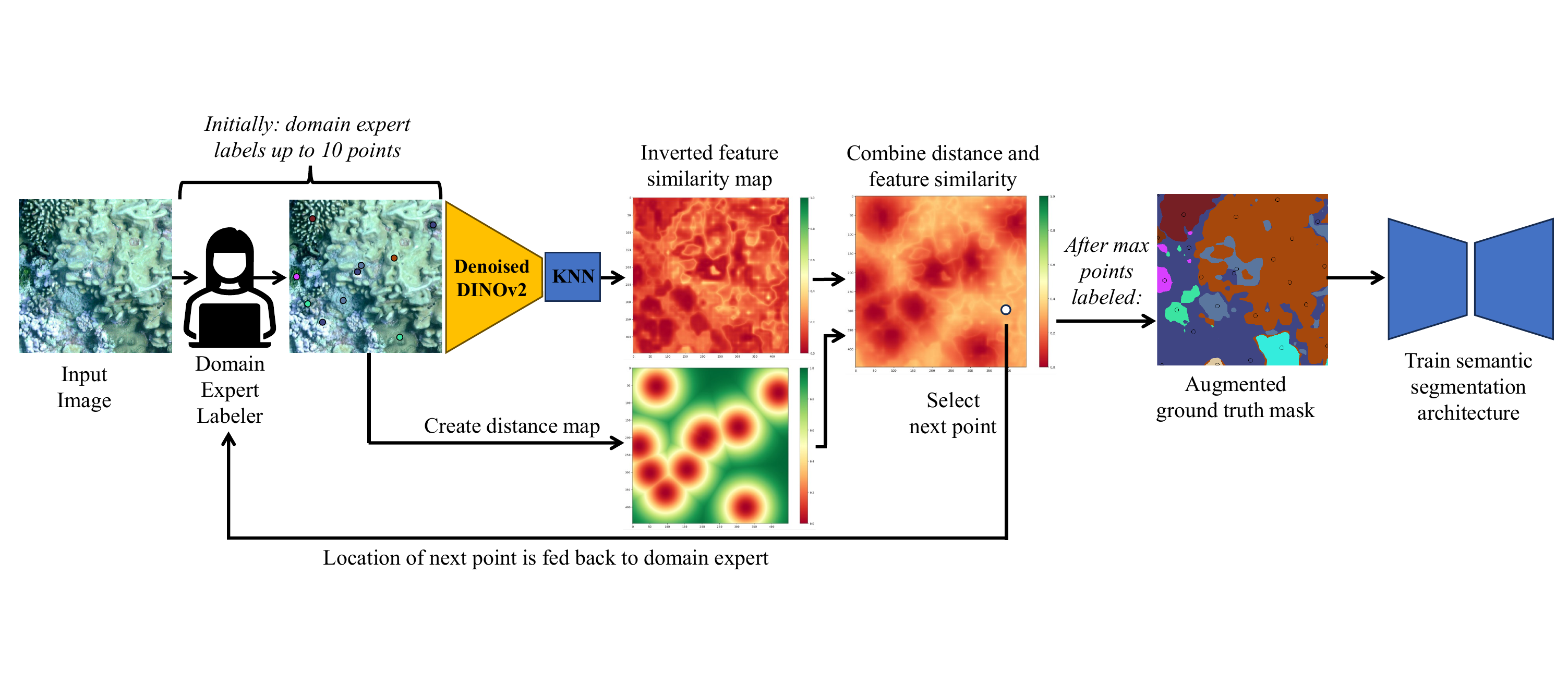}
\caption{Schematic of Proposed Human-in-the-Loop Labeling Approach. We combine domain expert knowledge with the model's internal uncertainty to improve point label selection. The process starts with inputting a coral image and having a marine scientist label up to 10 points centrally located on the largest instances (refer to Section~\ref{subsec:ablations} for an analysis of this value). A feature similarity map is then generated by computing cosine similarities between the labeled points and all other pixels. To promote exploration, we use a distance map in conjunction with the similarity map to create a combined probability mask for pixel selection. The chosen pixel is then sent back to the marine scientist for labeling, and the KNN is updated accordingly. Once the maximum number of points has been labeled, an augmented ground truth mask is created for use in training a semantic segmentation model. The semantic segmentation architecture is trained end-to-end on pairs of images and augmented ground truth masks (as seen in Fig.~\ref{fig:segpipeline}, which demonstrates that this model is then used to perform inference on new images).}
\label{fig:pipeline}
\end{figure*}

\subsection{Human-in-the-Loop Pixel Proposal}
\label{subsec:hil}
To enhance our propagation accuracy in cases with extremely sparse data, we design a novel labeling regime (Fig.~\ref{fig:pipeline}). Unlike previous methods that have utilized random or grid-spaced sparse point labels, we treat the point labeling task as a human-in-the-loop process. We assume the availability of a marine scientist who can collaboratively label a specific number of points, which will then be incorporated into our DINOv2 and KNN point label propagation approach.

To identify informative points for the marine scientist to annotate, we analyze the cosine similarity between the features of labeled and unlabeled pixels in the DINOv2 deep embedding space. Initially, we ask the marine scientist to label up to 10 pixels located centrally in the largest instances visible in the image (see Section~\ref{subsec:ablations} for an ablation study on this number). For instances requiring more than 10 labeled points, the human-in-the-loop regime iteratively proposes one point at a time for labeling, focusing on the areas of the image with the highest uncertainty. This uncertainty is quantified as the cosine similarity to the nearest labeled pixel.

To implement this, we apply the method described in Section~\ref{subsec:methodoverview} to obtain, upsample, and normalize the per-pixel feature embeddings. We then calculate a cosine similarity map (\ref{eq:cosine}) between the initially labeled pixels and every other pixel in the image. By inverting this map, we assign a higher probability of selection to pixel locations that exhibit low cosine similarity to the nearest labeled pixel:

\begin{equation}
    \label{eq:similarity}
    \mathrm{C}(x,y) = 1 - \max_{l \in \{1,\dots,L\}} \mathrm{sim(}\textbf{v}_{q}, {\textbf{v}}_l),
\end{equation}
where $\textbf{v}_q$ is the feature vector at location $(x,y)$.

To promote exploring the entire image, we generate a probabilistic distance map based on the labeled pixels. We calculate the Euclidean distance transform on a binary mask that indicates the positions of the labeled pixels, where the initial count is $L=10$:

\begin{equation}
    \label{eq:distance}
    \mathrm{D}(x,y) = \min_{(x',y')\in X} \sqrt{(x - x')^2 + (y - y')^2},
\end{equation}
where $X=\{(x_1,y_1),\dots,(x_L,y_L)\}$, which stores the pixel coordinates of the labeled points.

We then perform Gaussian smoothing over the distance transform and tune the smoothing parameter $\sigma$ in the ablation study in Section \ref{subsec:ablations}: 
\begin{equation}
    \label{eq:gaussian}
    \mathrm{D_{smooth}}(x,y) = 1 - \exp{\Big(-\frac{\mathrm{D}(x,y)^2}{2 \sigma^2}\Big)}.
\end{equation}

We combine the probabilistic cosine similarity map with the distance map, and weight the two terms with $\lambda$ (see hyperparameter tuning in Section \ref{subsec:ablations}):

\begin{equation}
    \label{eq:combine}
    \mathrm{M}(x,y) = \frac{ \mathrm{D_{smooth}}(x,y) + \lambda \, \mathrm{C}(x,y) }{\lambda + 1}.
\end{equation}

From the combined map we identify the next pixel for annotation by the expert by selecting the location $(\hat{x}, \hat{y})=\argmax_{(x,y)} \mathrm{M}(x, y)$ corresponding to the highest selection probability in $\mathrm{M}$.

\subsection{Semantic Segmentation}
\label{subsec:segmentation}

After point label propagation (Stage One), we use the augmented ground truth masks to train a fully supervised model for semantic segmentation (Stage Two). The model chosen for Stage Two can be tailored to meet the computational and inference time needs for deployment. We evaluate the performance of various semantic segmentation architectures in Section~\ref{subsec:ablations}. We compare the DeepLabv3+ architecture~\cite{chen2018encoder} with ResNet50~\cite{he2016deep}, Inceptionv3~\cite{szegedy2016rethinking} and Mobilenetv2 backbones with UNet~\cite{ronneberger2015u} and LinkNet~\cite{chaurasia2017linknet} with ResNet50~\cite{he2016deep} backbone.  We train the semantic segmentation architectures using the augmented ground truth masks, as outlined in Section~\ref{subsec:Implementation}.

\section{Experimental Setup}
\label{sec:experimentalsetup}
In this section, we outline the details of our implementation in Section~\ref{subsec:Implementation}, describe the evaluation datasets in Section~\ref{subsec:datasets}, and define the evaluation metrics in Section~\ref{subsec:evaluationmetrics}.

\subsection{Implementation}
\label{subsec:Implementation}

\subsubsection{Stage One: Point Label Propagation}
All experiments in this work are completed on a Quadro RTX 6000 GPU, and we calculate point label propagation times in Table~\ref{tab:stage1} with respect to this GPU. We implement our presented approach with Python and PyTorch \cite{paszke2019pytorch}.  In addition, we employ the Faiss library to enable faster implementation of K-Nearest Neighbors on GPU \cite{johnson2019billion}.  The denoised DINOv2 model and implementation is from \cite{yang2024denoising}.

\subsubsection{Stage Two: Semantic Segmentation}
\label{subsubsec:Implementation-seg}
We evaluate the propagated ground truth masks by training a model for semantic segmentation. We use TensorFlow to train a DeepLabv3+ model~\cite{chen2018encoder} with ResNet50 backbone~\cite{he2016deep}. We also compare alternate semantic segmentation architectures and backbones in the ablation study in Section~\ref{subsec:ablations}, and we report the inference times of these architectures in frames per second (fps, a higher number indicates faster inference), when inference is performed on a NVIDIA A100 GPU. During training we apply data augmentation techniques, including random horizontal and vertical flipping, and adjust the gain and gamma values within the range of $0.8 - 1.2$. The training process spans 500 epochs using the Adam optimizer with a learning rate of $0.001$. We report results on the test dataset based on the best epoch. 

Unlike prior works \cite{alonso2019coralseg} and \cite{raine2022point}, for Stage Two we include all classes for training and evaluation.  Previously, prior works have excluded the ``unknown" class in the UCSD Mosaics dataset (Section~\ref{subsec:datasets}) from training and test.  In this work, due to the sparsity of the labels considered in this setting, we include this class during training and test to ensure a fair evaluation of all models. If the ``unknown" class is excluded, this results in model under-fitting and misrepresentation of the model performance.

\subsection{Datasets}
\label{subsec:datasets}

\begin{figure}[t]
\centering
\includegraphics[width=0.7\textwidth, clip, trim=0cm 6cm 0cm 3cm]{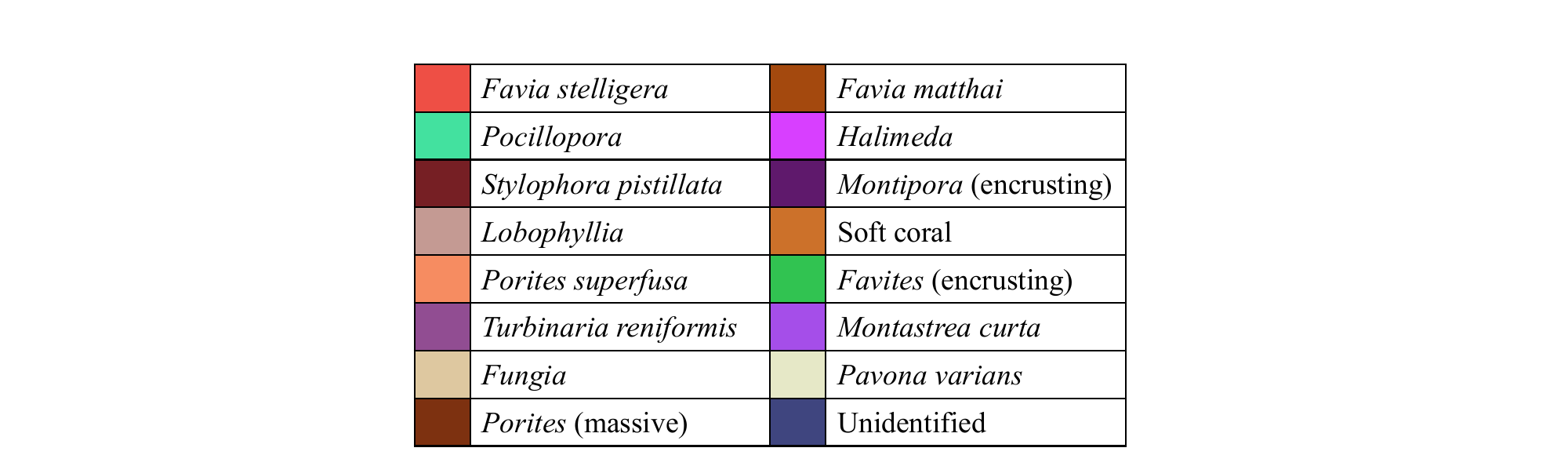}
\caption{Legend for the UCSD Mosaics dataset~\cite{edwards2017large, alonso2019coralseg}. This table displays the color coding for the semantic classes referenced throughout the paper. The legend corresponds to the classes present in the ground truth masks, point label propagation and segmentation outputs shown in Figs.~\ref{fig:frontpage}-\ref{fig:pipeline}, \ref{fig:ucsd} and  \ref{fig:qualitativeresults}-\ref{fig:denoise}. Only classes appearing in the visualizations included in this paper are shown.}
\label{fig:legend}
\end{figure}

Recent works have contributed datasets of underwater imagery for semantic segmentation~\cite{ziqiang2023coralvos, furtado2023deolhonoscorais, islam2020semantic}, however these datasets do not include multiple species of corals.  As coral imagery is highly complex, with overlapping instances and poorly defined outlines, it is extremely time-consuming and expensive for marine scientists to provide pixel-wise dense annotations to accompany the training images.  Therefore, the vast majority of multi-species coral datasets are labeled only with sparse point labels \eg~\cite{gonzalez2019seaview, chen2021new}. The UCSD Mosaics dataset is the only multi-species coral dataset with accompanying pixel-wise ground truth segmentation masks~\cite{alonso2019coralseg, raine2022point}.  We acknowledge that this is a significant limitation and challenge for the field of multi-species segmentation of coral imagery, and there is a clear need for more multi-species, densely labeled datasets of coral imagery fore comparison of approaches. In this work, we therefore use the UCSD Mosaics dataset~\cite{edwards2017large, alonso2019coralseg} for evaluation of our proposed approach and comparison to prior works. The ground truth masks are needed to accurately evaluate the performance of semantic segmentation models. The colorised legend for the pixel-wise ground truth segmentation masks is in Fig.~\ref{fig:legend} -- we refer to this legend throughout this paper when showing results on this dataset.

In \cite{alonso2019coralseg}, the authors provide a version of the dataset where the large orthomosaics have been saved as smaller images, however we have identified some corrupted ground truth masks and have excluded them (this resulted in 219 being removed from the training set, resulting in 3,974 images, and 32 were removed from the test set, leaving 696 images; examples of the corruption can be seen in Fig.~\ref{fig:ucsd}). We include a list of the corrupted filenames alongside our publicly available code\footnote{\url{https://github.com/sgraine/HIL-coral-segmentation}}. Each image measures 512 by 512 pixels and features 33 types of corals along with an class called `Unidentified'. To maintain consistency with \cite{alonso2019coralseg} and \cite{raine2022point}, we disregard this class during evaluation of the point label propagation task. We imitate the role of the domain expert in our human-in-the-loop labeling framework by obtaining the ground truth point label corresponding to the proposed location. For the first 10 pixels, we select the central pixel of the largest instances in the mask.

\subsection{Evaluation Metrics}
\label{subsec:evaluationmetrics}
We employ three commonly used metrics \cite{alonso2019coralseg,garcia2018survey,raine2022point} to evaluate the performance of our method against previous approaches:
\begin{enumerate}
\item Pixel Accuracy (PA), which measures the proportion of correctly predicted pixels out of all predicted pixels,
\item Mean Pixel Accuracy (mPA), the average pixel accuracy across all classes, and
\item Mean Intersection over Union (mIoU), representing the average of the IoU scores for each class.
\end{enumerate}

For all these metrics, a higher value signifies higher performance.  These metrics are used to quantify the performance of both Stage One and Stage Two. 

\newcommand{\scaleCorruptSets}{0.15\textwidth}
\begin{figure}
    \centering
    \setlength{\tabcolsep}{1pt}
    \centerline{\begin{tabular}{ccc}
     
    \includegraphics[width=\scaleCorruptSets]{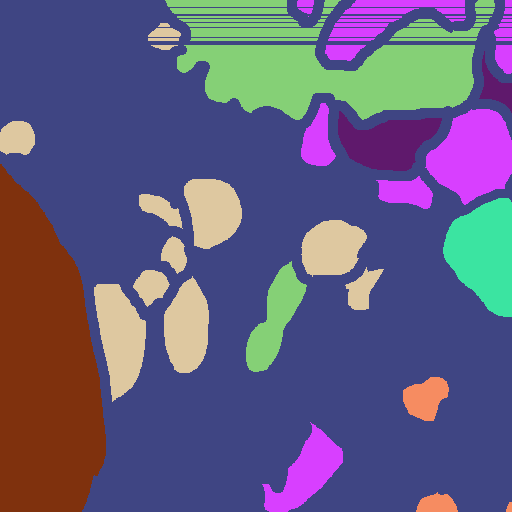} &
    \includegraphics[width=\scaleCorruptSets]{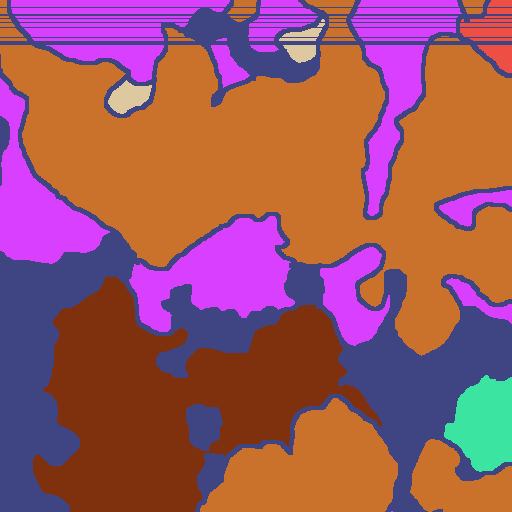} &
    \includegraphics[width=\scaleCorruptSets]{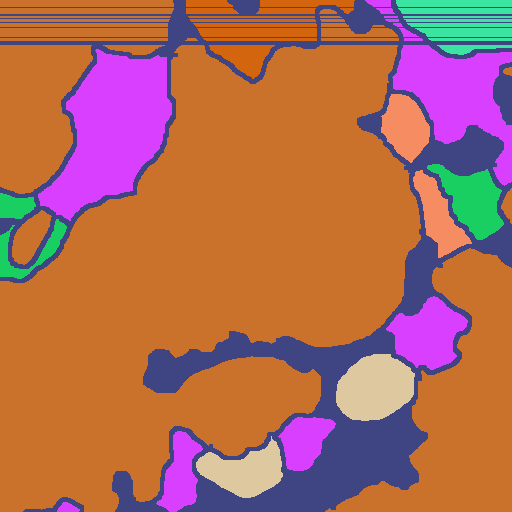} \\    

    \end{tabular}}
    \caption{Some of the ground truth masks in the UCSD Mosaics dataset showed signs of corruption, as illustrated at the top of these examples. The corresponding images and ground truth masks were removed from the training and test datasets. Refer to Fig.~\ref{fig:legend} for the color legend.}
    \label{fig:ucsd}
\end{figure}

\section{Results}
We compare our proposed approach to the point label propagation state-of-the-art in Section~\ref{subsec:sota} and then Section~\ref{subsec:ablations} provides various ablation studies. We describe our findings on the effect of point label quantity in Section~\ref{subsec:results-label-quantity} and the effect of point label placement style in Section~\ref{subsec:results-label-placement}.

\subsection{Comparison to State-of-the-art Methods}
\label{subsec:sota}

\subsubsection{Stage One: Point Label Propagation}
\label{subsubsec:pointprop}
In this section, we evaluate the performance of our approach and compare against a number of state-of-the-art methods, including: Fast Multi-level Superpixel Segmentation (\textit{Fast MSS}) ~\cite{pierce2020reducing}, a faster re-implementation of CoralSeg~\cite{alonso2019coralseg}, and Point Label Aware Superpixels~\cite{raine2022point}. In the case of~\cite{raine2022point}, we compare against the single method (\textit{Single}) and also the ensemble of three Point Label Aware algorithms (\textit{Ensemble}).

\begin{figure*}[t]
    \centering
    \centerline{\includegraphics[width=1.01\linewidth, clip, trim=0cm 4.5cm 0cm 0cm]{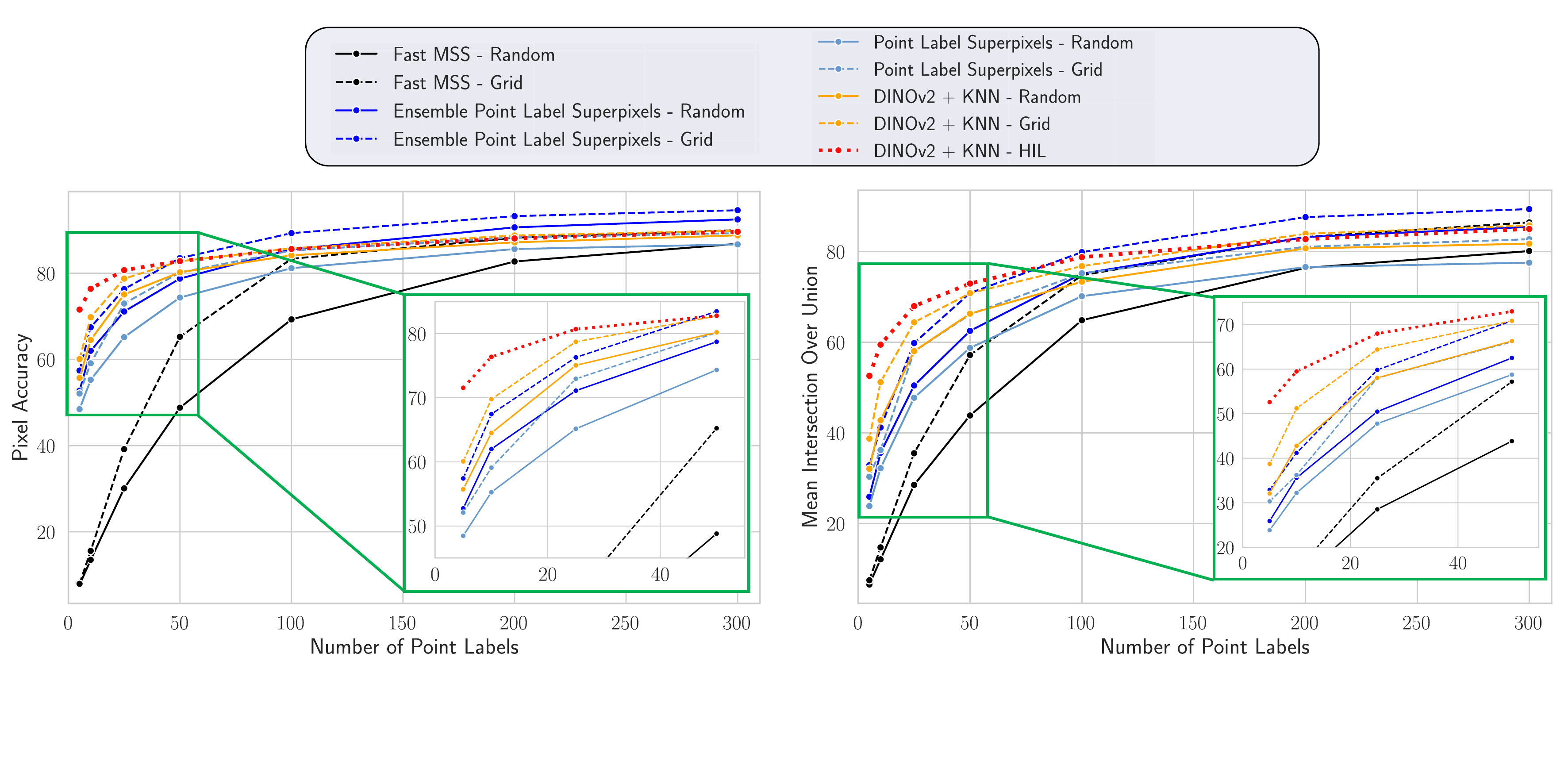}}
    \caption{Point Label Propagation Performance (Pixel Accuracy and Mean IoU): In the graphs, our proposed DINOv2 and KNN method is represented in orange for both random and grid labeling, while the red line shows the performance of the DINOv2 and KNN with the Human-in-the-Loop labeling approach. Our method notably surpasses previous approaches when only a small number of point labels are available, specifically between 5 and 25 points. However, when the number of points increases to 300, the performance of all approaches tends to converge.}
    \label{fig:point-graph}
\end{figure*}

As demonstrated in Table~\ref{tab:stage1}, using K-Nearest Neighbors with features extracted by the denoised DINOv2 foundation model~\cite{yang2024denoising} for point label propagation surpasses previous methods when dealing with a small number of point labels (5, 10, and 25 per image). When using our human-in-the-loop labeling framework with five point labels, the mIoU increases by 46.1\% and 22.6\% compared to the Fast MSS (F-MSS) and Point Label Aware Superpixel (PLAS) algorithms, respectively (Fig.~\ref{fig:point-graph}). Additionally, we see an improvement of 64.3\% and 17.3\% in pixel accuracy (Fig.~\ref{fig:point-graph}). Without the human-in-the-loop labeling framework, our method still exceeds the state-of-the-art PLAS by 3.5\% in pixel accuracy and 5.7\% in mIoU (if 5 grid points are used).

In a scenario not targeted by this paper, where up to 300 point labels are available, our approach shows performance comparable to single classifier methods. However, the ensemble of three Point Label Aware Superpixel classifiers outperforms our approach (Table~\ref{tab:stage1}).

\begin{table*}
\caption{Performance of Stage One: Point Label Propagation Approaches (Refer to Section~\ref{subsec:evaluationmetrics} for Metric Definitions), for 5~/~10~/~25~/~300 Point Labels. `F-MSS' is Fast MSS~\cite{pierce2020reducing}, `PLAS' is Point Label Aware Superpixels~\cite{raine2022point}, and `D+NN' is KNN with Denoised DINOv2~\cite{yang2024denoising} (Ours).}
\vspace*{-0.2cm}
\setlength{\tabcolsep}{3pt}

\centering
\scriptsize
\centerline{\begin{tabular}{@{}lcccccc}

\toprule
 & \textbf{Label} & \textbf{PA} & \textbf{mPA} & \textbf{mIoU}  & \textbf{Time per Image (s)} \\

\textbf{Method} & \textbf{Style} & \textbf{5 / 10 / 25 / 300} & \textbf{5 / 10 / 25 / 300} & \textbf{5 / 10 / 25 / 300} & \textbf{5 / 10 / 25 / 300} \\

\midrule
F-MSS & Rand. & 7.29 / 13.49 / 30.09 / 86.81 &  6.60 / 12.34 / 29.26 / 82.70 & 6.55 / 12.11 / 28.53 / 80.12 & 2.14 / 2.19 / 2.21 / 2.76 \\
F-MSS & Grid & 7.94 / 15.58 / 39.18 / 89.98 & 7.54 / 14.96 / 36.72 / 88.17 & 7.50 / 14.74 / 35.51 / \textit{86.44} & 2.43 / 2.45 / 2.36 / 2.96 \\
PLAS \textit{- Single} & Rand. & 48.45 / 55.26 / 65.16 / 86.68 & 32.03 / 41.44 / 57.65 / 81.74 & 23.86 / 32.22 / 47.76 / 77.56 & \textit{1.71} / \textit{2.00} / \textit{2.17} / \textit{1.93} \\
PLAS \textit{- Single} & Grid & 52.08 / 59.09 / 72.96 / 89.28 & 39.89 / 44.88 / 64.91 / 86.16 & 30.32 / 36.22 / 58.00 / 82.73 & \textbf{1.55} / \textbf{1.80} / \textbf{2.06} / \textbf{1.81} \\
PLAS \textit{- Ens.}  & Rand. & 52.73 / 62.00 / 71.11 / \textit{92.47}& 36.48 / 49.04 / 63.21 / \textit{89.93}& 25.91 / 35.6 / 50.46 / 85.45 & 4.27 / 4.55 / 5.02 / 5.35\\
PLAS \textit{- Ens.} & Grid & 57.41 / 67.45 / 76.31 / \textbf{94.60} & 44.36 / 54.44 / 69.13 / \textbf{92.49} & 32.89 / 41.19 / 59.82 / \textbf{89.38} & 4.06 / 4.25 / 5.15 / 5.28 \\
D+NN (Ours) & Rand. & 55.72 / 64.51 / 75.07 / 88.77 & 39.94 / 50.91 / 65.80 / 83.84& 32.09 / 42.79 / 58.04 / 81.75 & 4.88 / 4.55 / 4.74 / 4.90\\
D+NN (Ours) & Grid & \textit{60.08} / \textit{69.79} / \textit{78.74} / 89.86 & \textit{47.85} / \textit{58.39} / \textit{70.05} / 87.41 & \textit{38.72} / \textit{51.20} / \textit{64.40} / 85.77 & 4.78 / 4.70 / 4.79 / 4.69 \\
D+NN (Ours) & HIL & \textbf{71.56} / \textbf{76.38} / \textbf{81.27} / 89.61 & \textbf{61.46} / \textbf{69.87} / \textbf{75.91} / 86.45 & \textbf{52.60} / \textbf{59.48} / \textbf{67.97} / 85.00 & 4.74 / 4.98 / 20.0 / 273.08 \\
\bottomrule
\end{tabular}}
\vspace*{-0.2cm}
\label{tab:stage1}
\end{table*}

Our approach, which combines DINOv2 with our human-in-the-loop labeling regime, shows comparable computation times per image to the Point Label Aware Superpixel ensemble for small numbers of point labels (Table~\ref{tab:stage1}). However, when applied to a large number of points \ie 300 points, the computation time becomes excessively high due to the clustering in the deep feature space required to generate the feature similarity map at each iteration (\ref{eq:similarity}). If there are a large number of point labels available, it is more suitable to use the grid-spaced version. We emphasize that the primary use case for human-in-the-loop labeling is to enhance performance in scenarios with very sparse point labels (5-25 points).

Fig.~\ref{fig:qualitativeresults} showcases outputs from our proposed method, illustrating it produces pixel-wise augmented ground truth masks which largely agree with the provided ground truth, even if there are very few sparse point labels available. This figure emphasizes that grid-based sparse labels offer better coverage compared to randomly placed sparse labels. For instance, in row 6, the Fast MSS approach~\cite{pierce2020reducing} misses an entire beige segment with randomly placed points, while the grid points capture it.

\newcommand{\scaleMaskSets}{0.14\textwidth}
\renewcommand{\arraystretch}{0.7} %
\begin{figure*}
    \centering
    \small
    \setlength{\tabcolsep}{1pt}
    \centerline{\begin{tabular}{cccccccc}
      & &  & \footnotesize{\textbf{Fast MSS}} & \footnotesize{\textbf{Fast MSS}} & \footnotesize{\textbf{PLAS}} & \footnotesize{\textbf{PLAS}} & \footnotesize{\textbf{Ours}} \\
      & \footnotesize{\textbf{Query Image}} & \footnotesize{\textbf{Ground Truth}} & \footnotesize{\textbf{Random Points}} & \footnotesize{\textbf{Grid Points}} & \footnotesize{\textbf{Random Points}} & \footnotesize{\textbf{Grid Points}} & \footnotesize{\textbf{HIL Points}} \\

    \rotatebox{90}{\hspace{0.7cm}\footnotesize{\textbf{5 Labels}}} & \includegraphics[width=\scaleMaskSets]{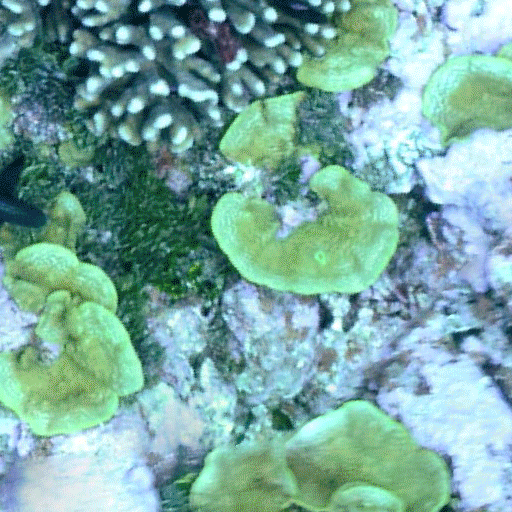} &
    \includegraphics[width=\scaleMaskSets]{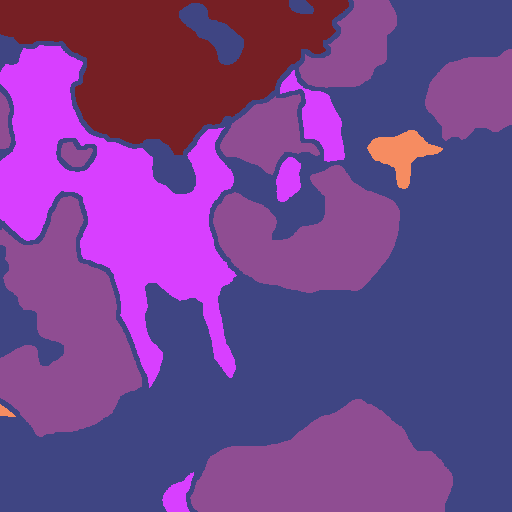} &
    \includegraphics[width=\scaleMaskSets]{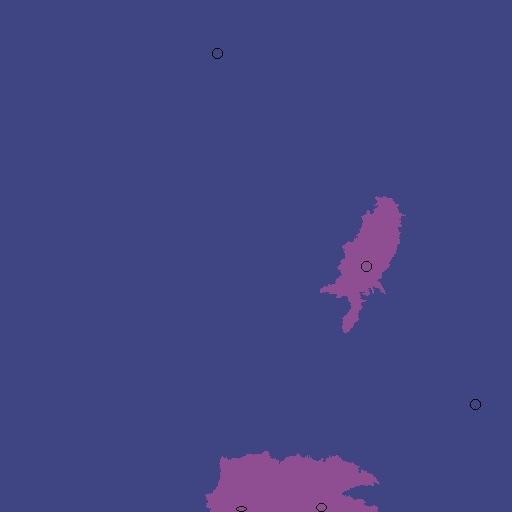} &
    \includegraphics[width=\scaleMaskSets]{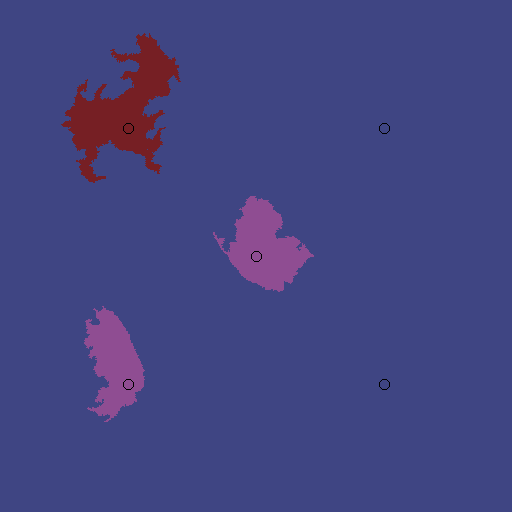} &
    \includegraphics[width=\scaleMaskSets]{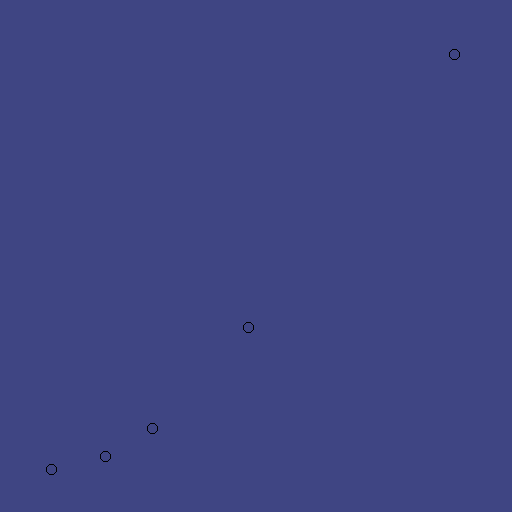} &
    \includegraphics[width=\scaleMaskSets]{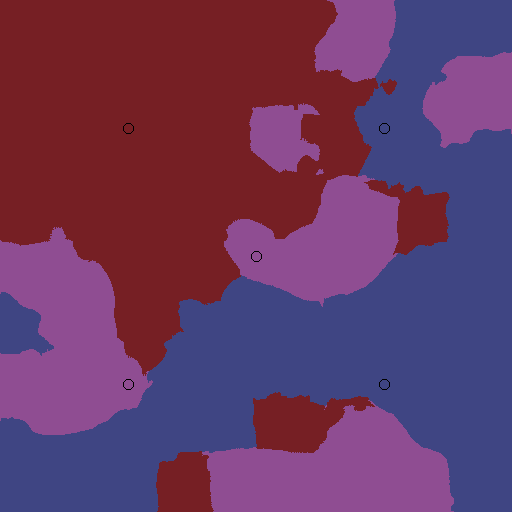} &
    \includegraphics[width=\scaleMaskSets]{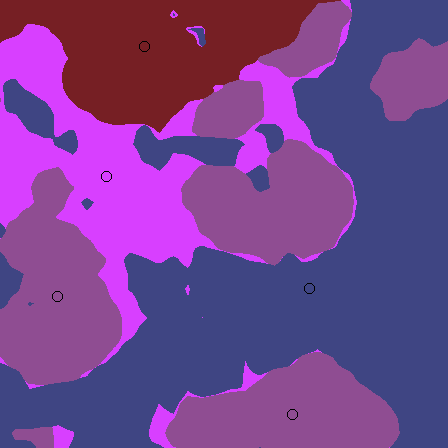} \\    

    \rotatebox{90}{\hspace{0.7cm}\footnotesize{\textbf{5 Labels}}} & \includegraphics[width=\scaleMaskSets]{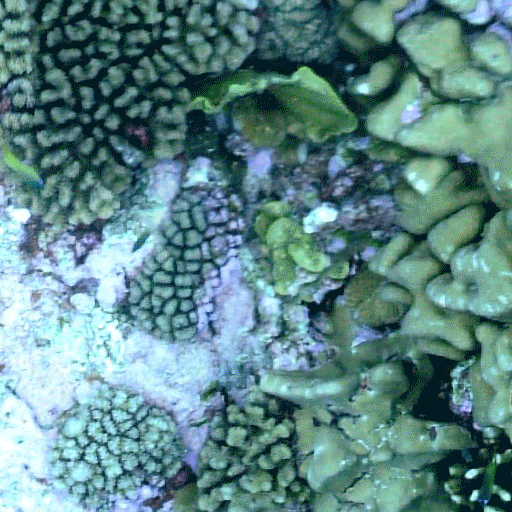} &
    \includegraphics[width=\scaleMaskSets]{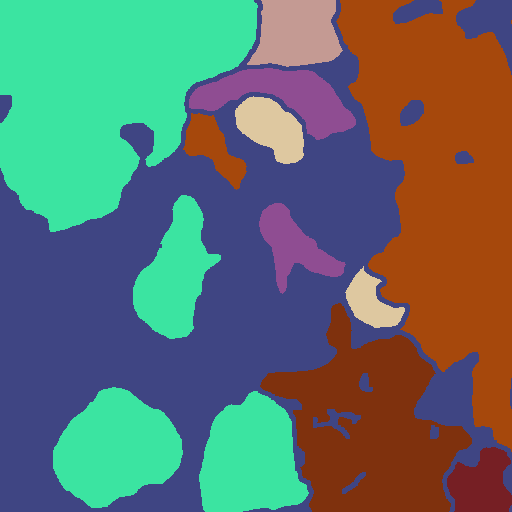} &
    \includegraphics[width=\scaleMaskSets]{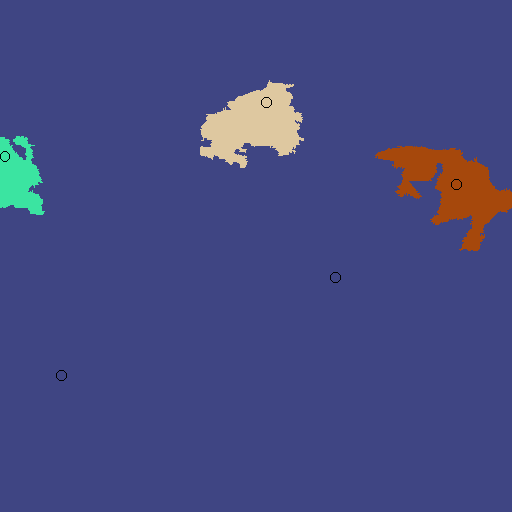} &
    \includegraphics[width=\scaleMaskSets]{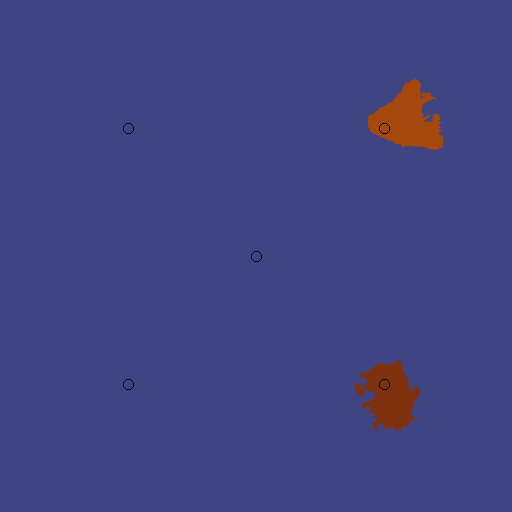} &
    \includegraphics[width=\scaleMaskSets]{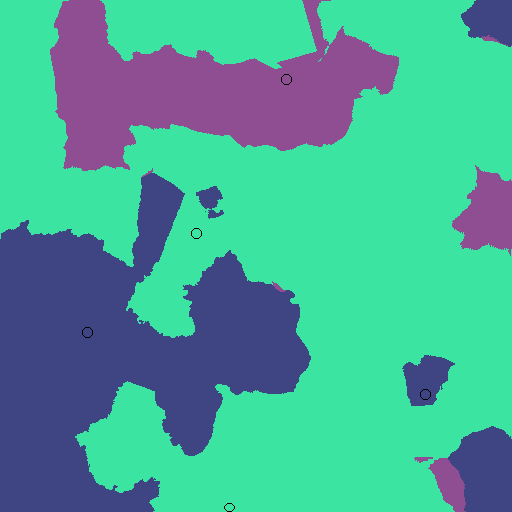} & 
    \includegraphics[width=\scaleMaskSets]{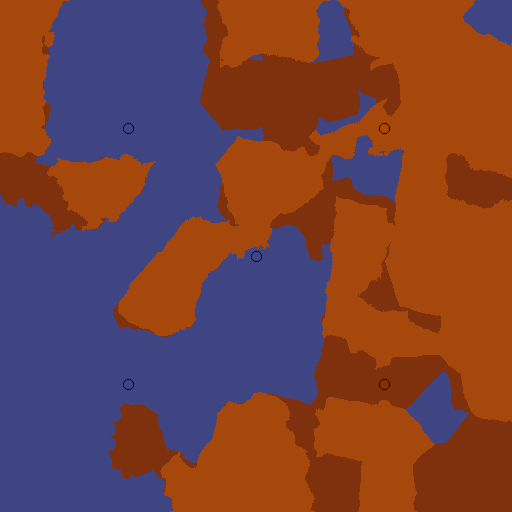} &
    \includegraphics[width=\scaleMaskSets]{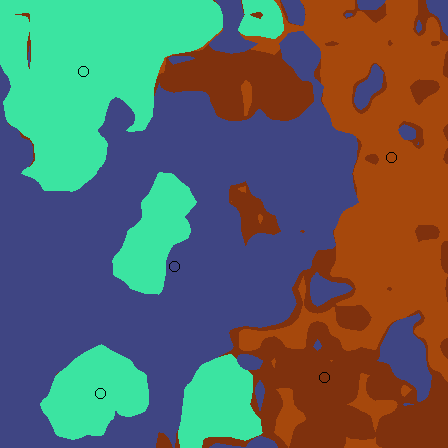} \\    

    \rotatebox{90}{\hspace{0.7cm}\footnotesize{\textbf{5 Labels}}} & \includegraphics[width=\scaleMaskSets]{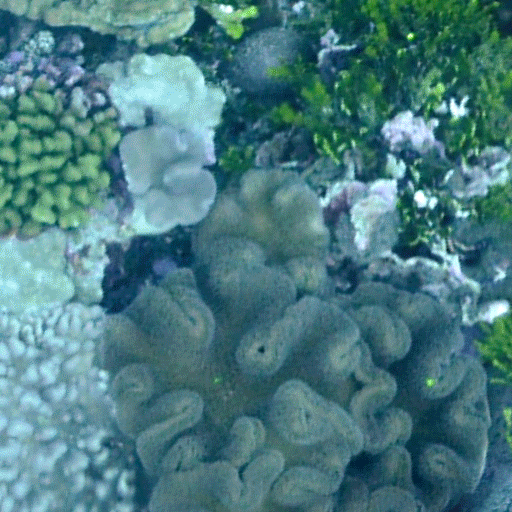} &
    \includegraphics[width=\scaleMaskSets]{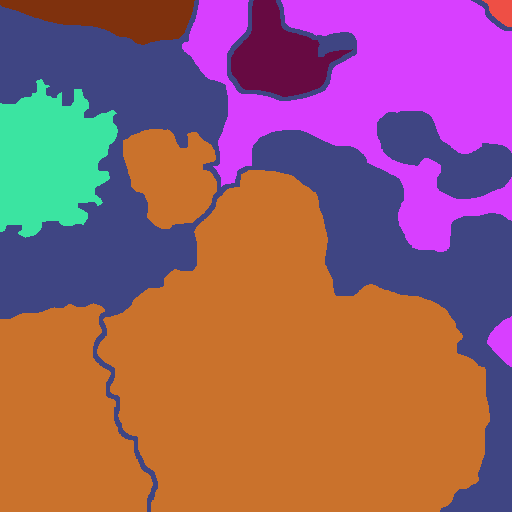} &
    \includegraphics[width=\scaleMaskSets]{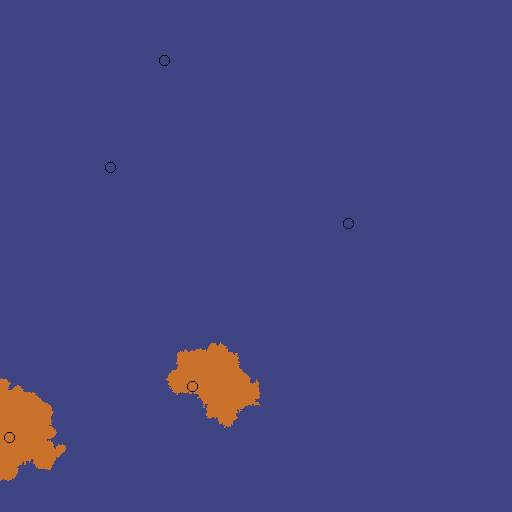} &
    \includegraphics[width=\scaleMaskSets]{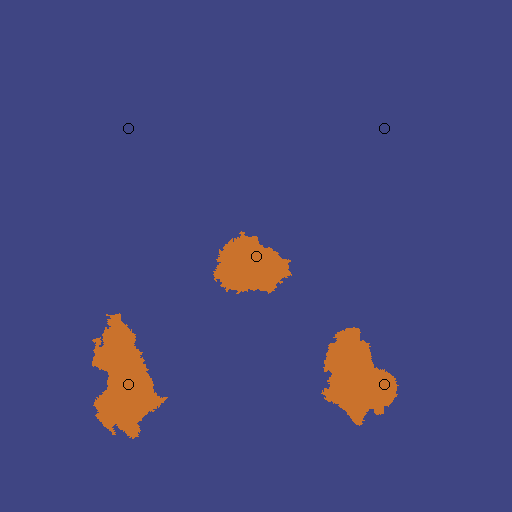} &
    \includegraphics[width=\scaleMaskSets]{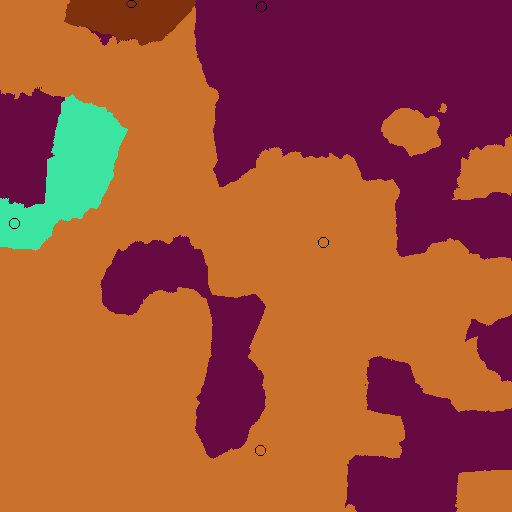} & 
    \includegraphics[width=\scaleMaskSets]{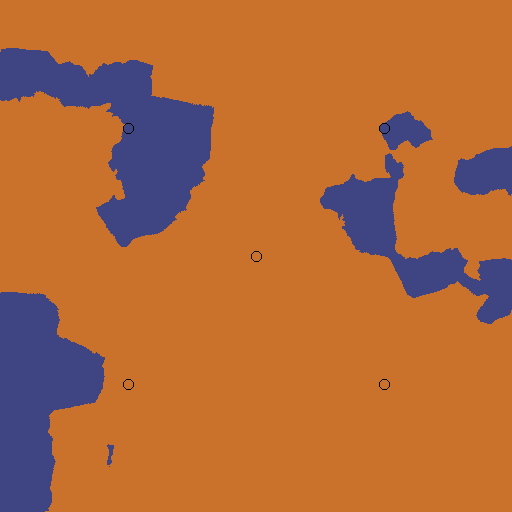} &
    \includegraphics[width=\scaleMaskSets]{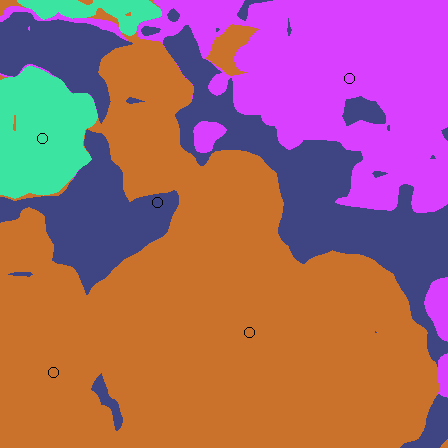} \\  

    \rotatebox{90}{\hspace{0.7cm}\footnotesize{\textbf{5 Labels}}} & \includegraphics[width=\scaleMaskSets]{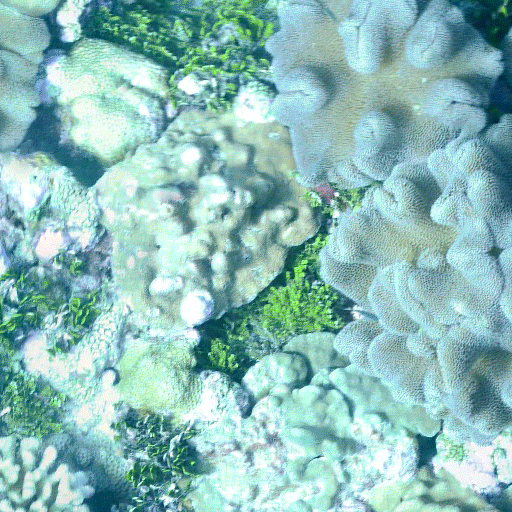} &
    \includegraphics[width=\scaleMaskSets]{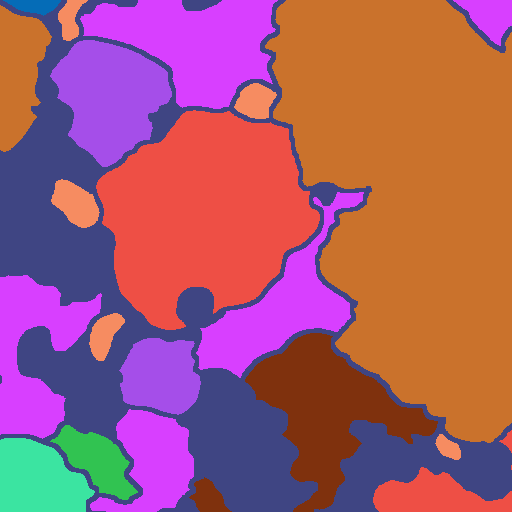} &
    \includegraphics[width=\scaleMaskSets]{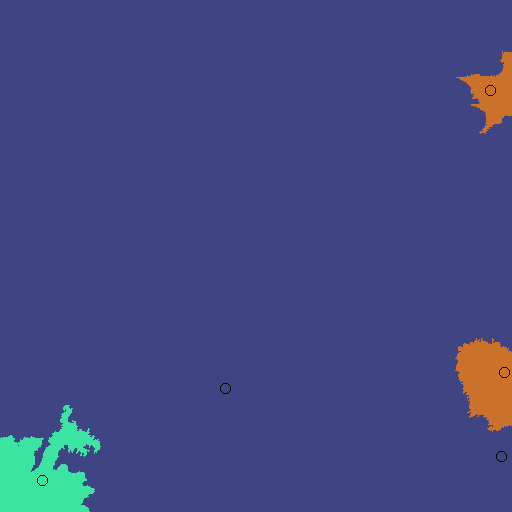} &
    \includegraphics[width=\scaleMaskSets]{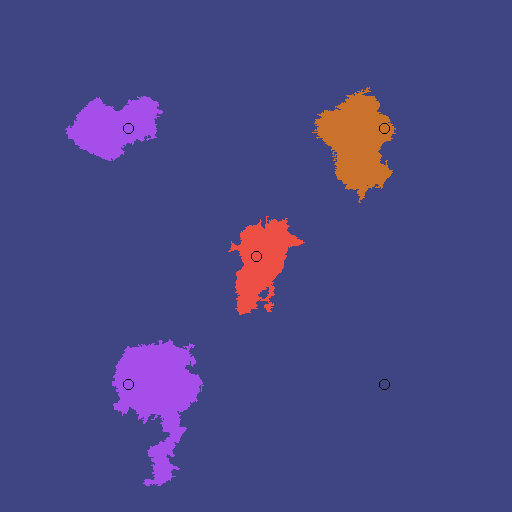} &
    \includegraphics[width=\scaleMaskSets]{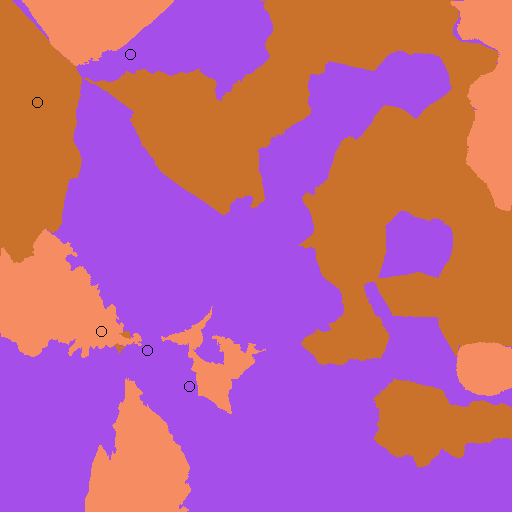} & 
    \includegraphics[width=\scaleMaskSets]{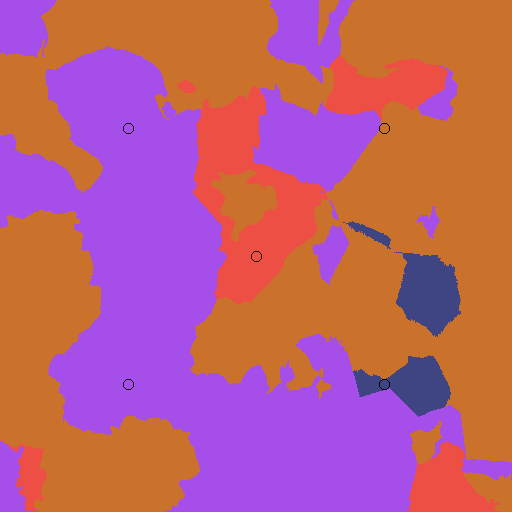} &
    \includegraphics[width=\scaleMaskSets]{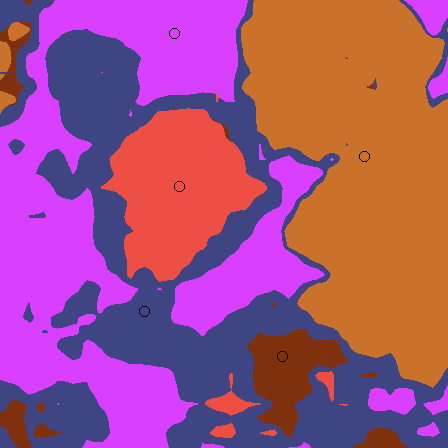}  \\
    \midrule

    \rotatebox{90}{\hspace{0.55cm}\footnotesize{\textbf{300 Labels}}} & \includegraphics[width=\scaleMaskSets]{Figures/example-masks/set1/FR9_6656_7168_8192_8704.png} &
    \includegraphics[width=\scaleMaskSets]{Figures/example-masks/set1/FR9_6656_7168_8192_8704_gt.png}  &
    \includegraphics[width=\scaleMaskSets]{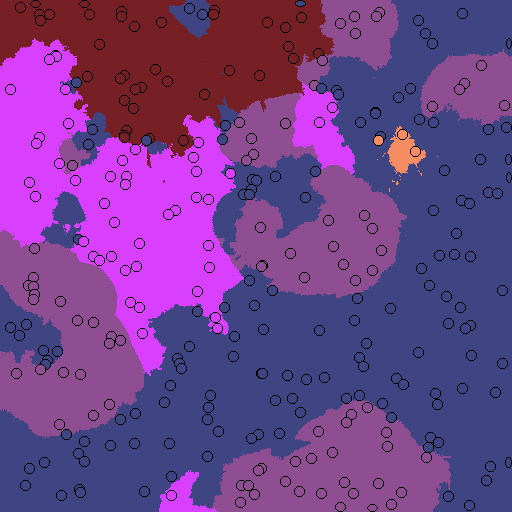} &
    \includegraphics[width=\scaleMaskSets]{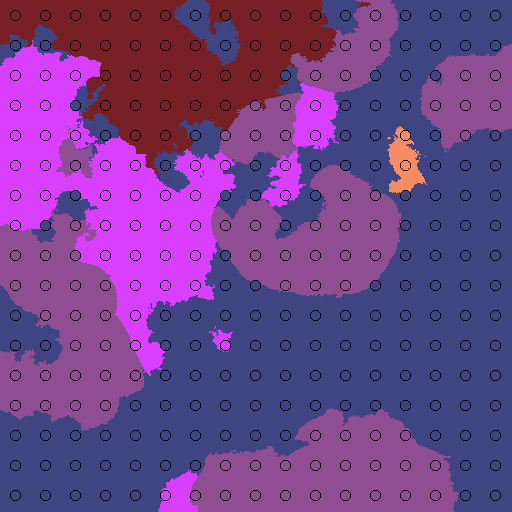} &
    \includegraphics[width=\scaleMaskSets]{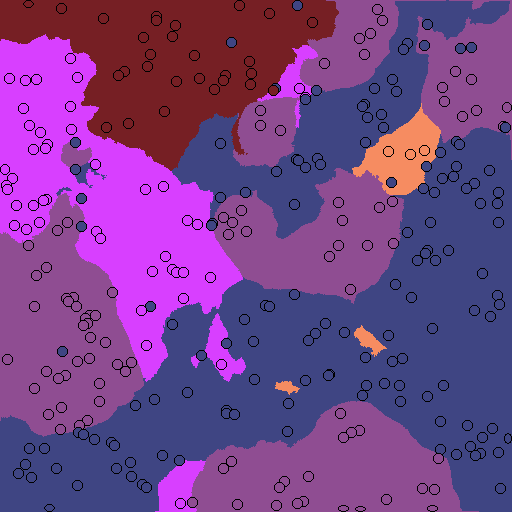} &
    \includegraphics[width=\scaleMaskSets]{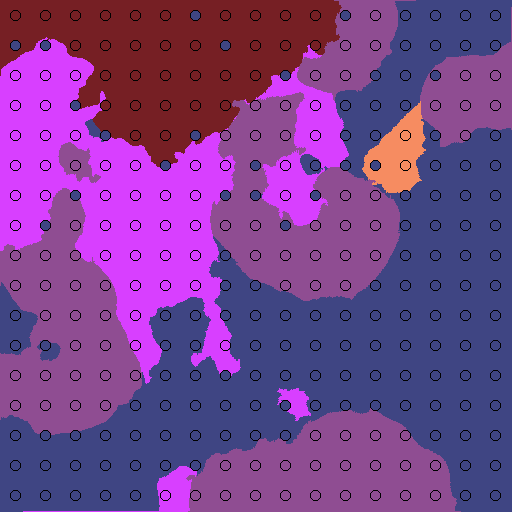} &
    \includegraphics[width=\scaleMaskSets]{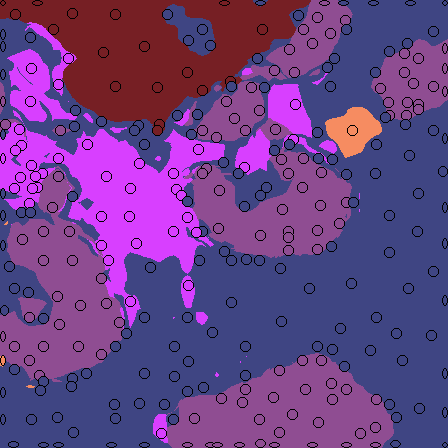} \\ 

     \rotatebox{90}{\hspace{0.55cm}\footnotesize{\textbf{300 Labels}}}  & \includegraphics[width=\scaleMaskSets]{Figures/example-masks/set2/FR9_8704_9216_4096_4608.png} &
    \includegraphics[width=\scaleMaskSets]{Figures/example-masks/set2/FR9_8704_9216_4096_4608_gt.png} &
    \includegraphics[width=\scaleMaskSets]{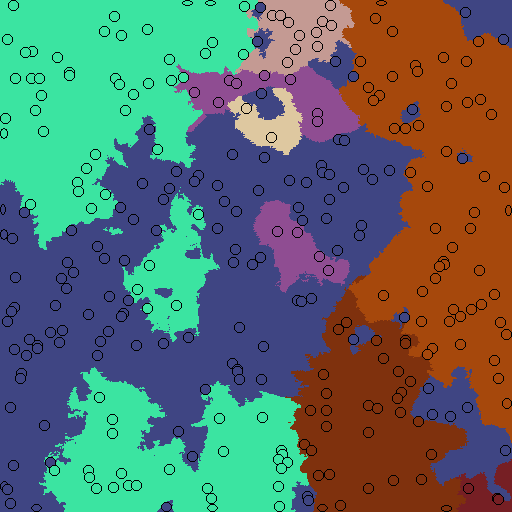} &
    \includegraphics[width=\scaleMaskSets]{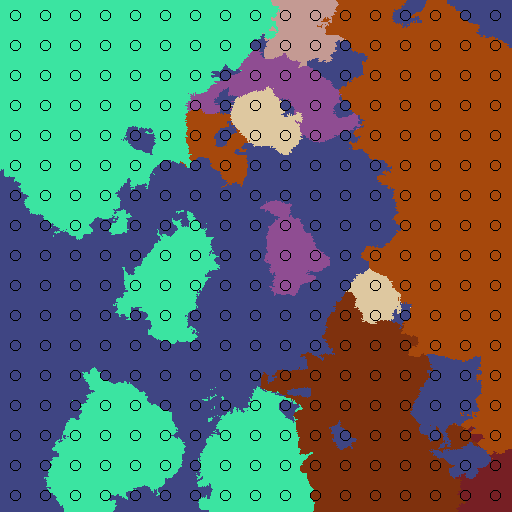} &
    \includegraphics[width=\scaleMaskSets]{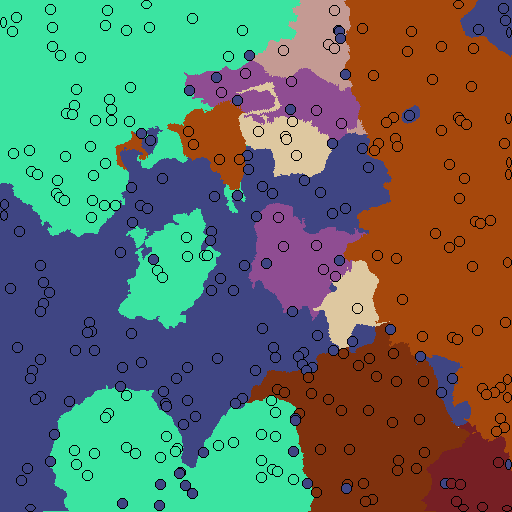} &
    \includegraphics[width=\scaleMaskSets]{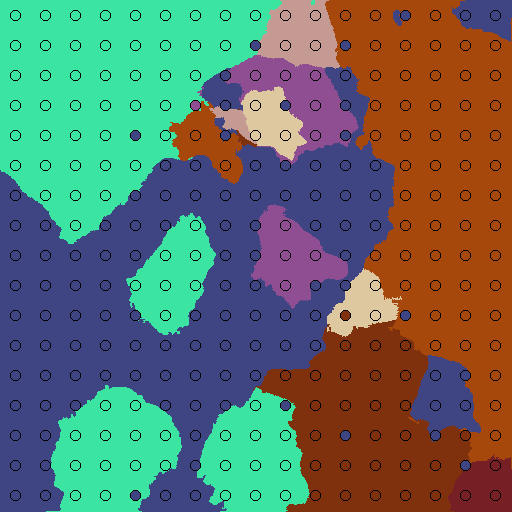} &
    \includegraphics[width=\scaleMaskSets]{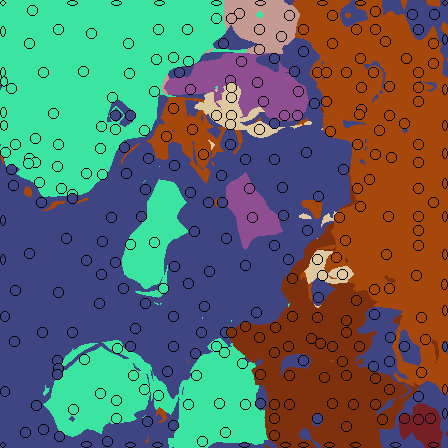} \\ 

    \rotatebox{90}{\hspace{0.55cm}\footnotesize{\textbf{300 Labels}}}  & \includegraphics[width=\scaleMaskSets]{Figures/example-masks/set3/FR7_1536_2048_2560_3072.png} &
    \includegraphics[width=\scaleMaskSets]{Figures/example-masks/set3/FR7_1536_2048_2560_3072_gt.png} &
    \includegraphics[width=\scaleMaskSets]{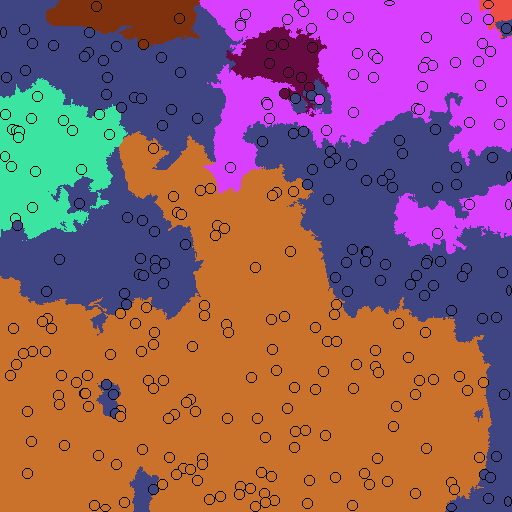} &
    \includegraphics[width=\scaleMaskSets]{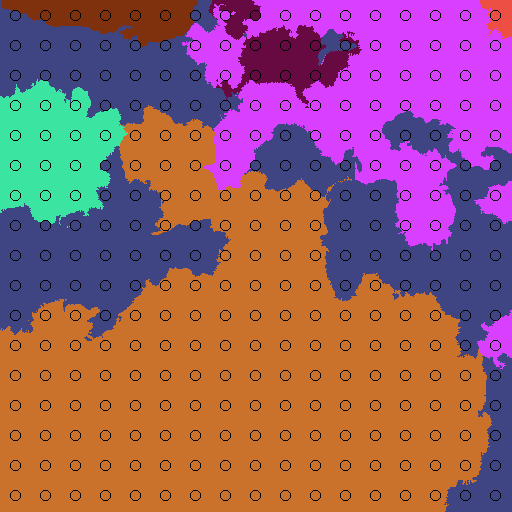} &
    \includegraphics[width=\scaleMaskSets]{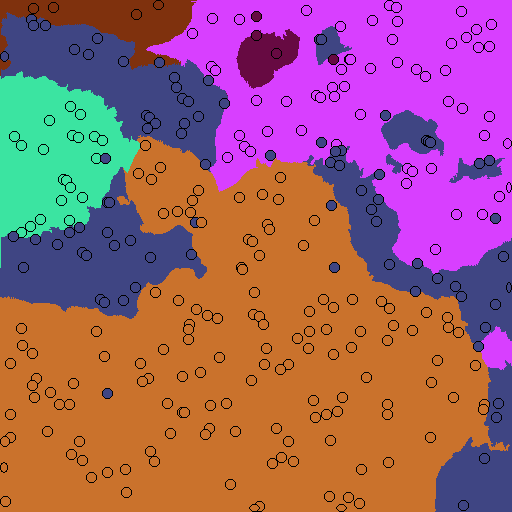} &
    \includegraphics[width=\scaleMaskSets]{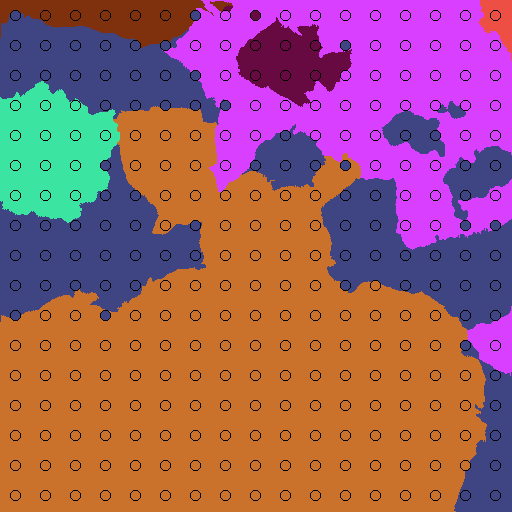} &
    \includegraphics[width=\scaleMaskSets]{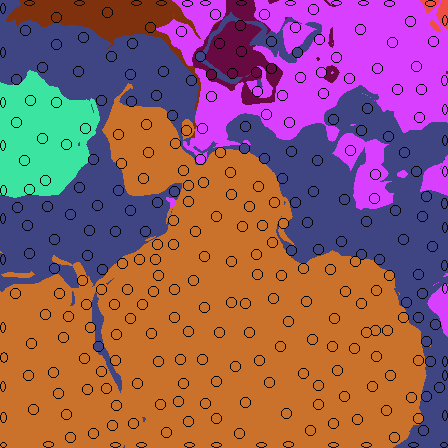} \\   

     \rotatebox{90}{\hspace{0.55cm}\footnotesize{\textbf{300 Labels}}}  & \includegraphics[width=\scaleMaskSets]{Figures/example-masks/set4/PALWave37_3072_3584_2560_3072.png} &
    \includegraphics[width=\scaleMaskSets]{Figures/example-masks/set4/PALWave37_3072_3584_2560_3072_gt.png} &
    \includegraphics[width=\scaleMaskSets]{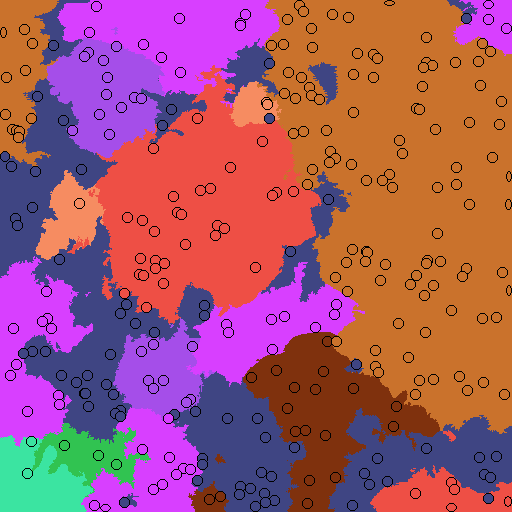} &
    \includegraphics[width=\scaleMaskSets]{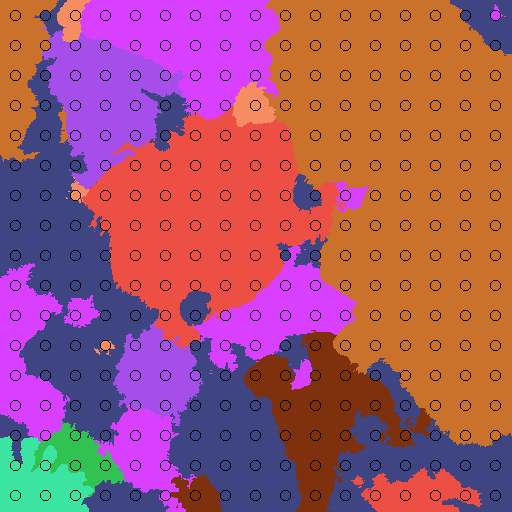} &
    \includegraphics[width=\scaleMaskSets]{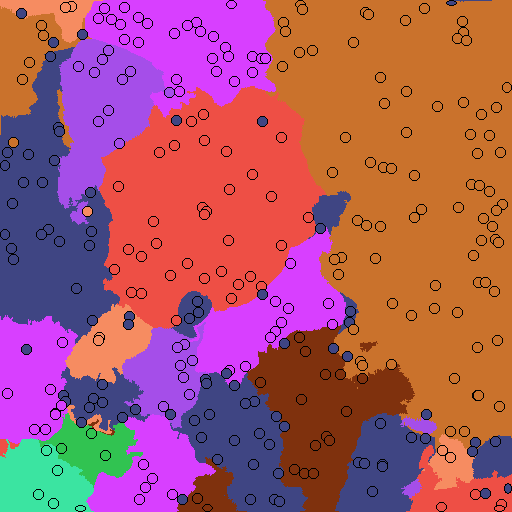} &
    \includegraphics[width=\scaleMaskSets]{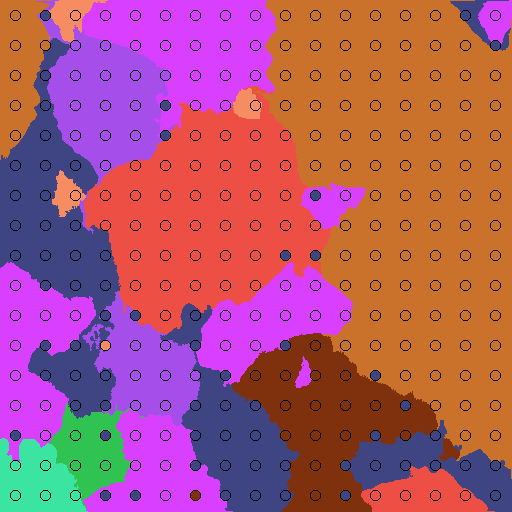} &
    \includegraphics[width=\scaleMaskSets]{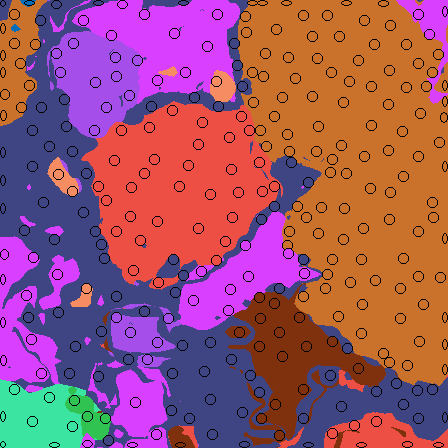} \\ 
    
\end{tabular}}
\vspace*{-0.2cm}
    \caption{\linespread{1.2}\selectfont A comparison is made between Fast MSS~\cite{pierce2020reducing},  Point Label Aware Superpixel (PLAS)~\cite{raine2022point}, and our approach, which combines denoised DINOv2 features~\cite{yang2024denoising}, K-Nearest Neighbors, and our Human-in-the-Loop labeling scheme. All methods are demonstrated on the same four examples. The top section shows point label propagation with 5 labels, while the bottom section displays propagation with 300 labels. Black circles indicate the point labels used for propagation within each output augmented mask. While all methods effectively propagate ground truth with 300 labels, the Fast MSS and PLAS methods struggle to produce usable augmented ground truth with only 5 labels. In contrast, our approach generates augmented ground truth that accurately resembles the ground truth even with the sparse 5 label setting. Refer to Fig.~\ref{fig:legend} for the color legend.}
    \label{fig:qualitativeresults}
\end{figure*}

The failure modes of Fast MSS and the Point Label Aware Superpixel approach in the case of 5 pixel labels are also evident in Fig.~\ref{fig:qualitativeresults}. Fast MSS does not generate useful segments because it only includes segments with point labels in the augmented ground truth mask. When segments lack any points (a common occurrence in this setting), they are assigned to the unknown/unlabeled class, resulting in the majority of the mask being this class. Conversely, the PLAS approach labels any superpixel segment without a point based on feature similarity with labeled segments, leading to significant over-prediction of classes. Additionally, PLAS requires sufficient points for the conflict loss function to ensure superpixel boundaries conform accurately to species~\cite{raine2022point}.

Our DINOv2 and KNN approach accurately generates augmented ground truth masks, even with extremely sparse labels. However, a limitation is that small species segments can be missed when there are very few point labels available. For example, in row 1 of Fig.~\ref{fig:qualitativeresults}, the orange segment is not included in the augmented ground truth. Future work could address this by incorporating mechanisms that emphasize smaller species segments and prevent the model from being biased towards larger instances.

\subsubsection{Stage Two: Semantic Segmentation}
\label{subsubsec:seg}

We trained a DeepLabv3+ model for semantic segmentation using the training regime and hyperparameters in Section~\ref{subsec:Implementation}, with results shown in Table~\ref{tab:stage2}. The performance improvements for  point label propagation are reflected when training a model on augmented ground truth masks. When trained on augmented ground truth masks generated using our DINOv2 and KNN approach with our human-in-the-loop labeling regime, the DeepLabv3+ model outperforms the previous state-of-the-art Point Label Aware Superpixels~\cite{raine2022point} by 8.8\% in pixel accuracy and by 13.5\% in mIoU with 5 point labels. Even without the human-in-the-loop labeling regime, our denoised DINOv2 and KNN approach still surpasses prior methods by 6.5\% in pixel accuracy and by 6.2\% in mIoU (5 point labels). Notably, the DeepLabv3+ model trained on our DINOv2 and KNN approach with grid-spaced labels slightly outperforms prior approaches in the 300 label case, despite not outperforming Point Label Aware Superpixels~\cite{raine2022point} on the point label propagation task. This suggests that once the first stage reaches a performance threshold, the second stage performance may saturate.

For the 10 and 25 point label settings, the DeepLabv3+ model trained on masks generated from grid labels exhibits higher pixel accuracy (77.6\% and 85.9\% for 10 and 25 points respectively) than the HIL generated masks (71.0\% and 81.7\%). The pixel accuracy calculates the overall correctly classified pixels, which does not take into consideration the class imbalance in the dataset. The mean pixel accuracy and mIoU are calculated as the average per-class scores, and better represent the performance across all classes: the HIL approach outperforms the grid approach by 10.0\% and 2.0\% for mean pixel accuracy and by 3.4\% and 2.1\% for mIoU for the 10 point and 25 point cases, respectively (Table~\ref{tab:stage2}). 

In these results (Table~\ref{tab:stage2}), we train and evaluate the models with the ``unknown" class included in the dataset. This class was omitted from training and test for the prior approaches \cite{raine2022point, alonso2019coralseg} because it can contain pixels which belong to one of the ``known" classes, thus introducing noise into the training signal.  In this case, we aim to establish the efficacy of the different point label propagation approaches by quantifying the performance of models trained on the masks.  For the extremely sparse label setting we consider in this work, baseline comparison methods generate propagated ground truth masks dominated by the unknown class (most notably the Fast-MSS approach, as seen in Fig.~\ref{fig:qualitativeresults}). When these ``unknown" pixels are not considered in segmentation task, it leads to model under-fitting which is then misrepresented by the metrics. This effect did not impact results for the original 100-300 point label setting considered in \cite{raine2022point, alonso2019coralseg}, as the larger number of points led to propagated ground truth masks which more closely resembled the ground truth.  

\begin{table*}
\caption{Performance of Stage Two: Semantic Segmentation with DeepLabv3+ (Refer to Section~\ref{subsec:evaluationmetrics} for Metric Definitions), for 5 / 10 / 25 / 300 Point Labels. `F-MSS' is Fast MSS~\cite{pierce2020reducing}, `PLAS' is Point Label Aware Superpixels~\cite{raine2022point}, and `D+NN' is KNN with Denoised DINOv2~\cite{yang2024denoising} (Ours). For all rows, the semantic segmentation architecture used is the Deeplabv3+ architecture~\cite{chen2018encoder} with ResNet50 backbone~\cite{he2016deep}.}
\vspace*{-0.2cm}
\centering
\footnotesize
\centerline{\begin{tabular}{@{}lccccc}

\toprule
 & \textbf{Label} & \textbf{PA} & \textbf{mPA} & \textbf{mIoU}  \\

\textbf{Method} & \textbf{Style} & \textbf{5 / 10 / 25 / 300} & \textbf{5 / 10 / 25 / 300} & \textbf{5 / 10 / 25 / 300} \\

\midrule
F-MSS~\cite{pierce2020reducing} & Grid & 48.05 / 51.06 / 57.41 / 85.34 & 5.97 / 10.39 / 14.83 / \textit{63.54} & 4.41 / 8.56 / 13.04 / 52.16 \\

PLAS \textit{- Ens.}~\cite{raine2022point} & Grid & 65.73 / 70.18 / 73.60 / 83.72 & 27.96 / 37.64 / 48.53 / \textbf{66.89} & 19.48 / 26.01 / 36.27 / 52.83 \\

D+NN (Ours) & Grid & \textit{72.24} / \textbf{77.64} / \textbf{85.93} / \textit{85.93} & \textit{32.16} / \textit{41.61} / \textit{52.59} / 62.99 & \textit{25.66} / \textit{34.80} / \textit{43.41} / \textit{54.07}  \\

D+NN (Ours) & HIL & \textbf{74.53} / \textit{71.04} / \textit{81.69} / \textbf{86.29} & \textbf{41.47} / \textbf{51.62} / \textbf{54.31} / 63.47 & \textbf{32.96} / \textbf{38.21} / \textbf{45.46} / \textbf{54.62} \\
\bottomrule
\end{tabular}}
\vspace*{-0.2cm}
\label{tab:stage2}
\end{table*}

\subsection{Ablation Study}
\label{subsec:ablations}

\subsubsection{Denoising DINOv2 Features}
\label{subsubsec:ablation-denoising}
We employ the denoised version of DINOv2 as described in~\cite{yang2024denoising} and highlight its effectiveness as a feature extractor through the results shown in Table~\ref{table:denoise}. We present a comparison between the original~\cite{oquab2024dinov} and denoised~\cite{yang2024denoising} DINOv2 deep feature embeddings in Fig.~\ref{fig:denoise}. Additionally, we compare the performance of DINOv2 trained with registers~\cite{darcet2023vision} to the denoised version trained with registers~\cite{darcet2023vision, yang2024denoising}, though this approach did not yield any improvement.  The denoising with registers approach slightly outperforms the denoising version for mean pixel accuracy only in the 5 and 10 point label cases, and for all other metrics and label quantities there is no improvement from the addition of registers.

\begin{table}[t]
\setlength{\tabcolsep}{4px}
\renewcommand{\arraystretch}{0.7} %
\caption{Effect of Different DINOv2 Feature Extractors (Refer to Section~\ref{subsec:evaluationmetrics} for Metric Definitions)} 
\label{table:denoise}
\centering
\begin{tabularx}{\columnwidth}{@{}>{\centering\arraybackslash}p{1.5cm}>{\centering\arraybackslash}p{1.5cm}>{\centering\arraybackslash}X>{\centering\arraybackslash}X>{\centering\arraybackslash}X}
\toprule
           &  & \textbf{PA} & \textbf{mPA} & \textbf{mIoU}\\
\textbf{Denoising} & \textbf{Registers} & \textbf{5 / 10 / 25 / 300} & \textbf{5 / 10 / 25 / 300} & \textbf{5 / 10 / 25 / 300}\\
\midrule

\crossmark & \crossmark & 68.58 / 73.32 / 76.94 / 88.10 & 60.23 / 68.04 / 70.97 / \textit{85.58} & 50.28 / 55.76 / 61.61 / \textit{83.79} \\
\crossmark & \checkmark & 68.49 / 73.12 / 76.65 / 87.41 & 59.79 / 67.48 / 72.44 / 84.84 & 49.80 / 55.96 / 61.46 / 82.68 \\
\checkmark & \checkmark & \textit{70.15} / \textit{75.41} / \textit{78.88} / \textit{88.16}  & \textbf{61.85} / \textbf{70.75} / \textit{75.81} / 85.42 & \textit{52.36} / \textit{59.47} / \textit{67.28} / 83.68 \\
\checkmark & \crossmark & \textbf{71.57} / \textbf{76.38} / \textbf{80.71} / \textbf{89.61} & \textit{61.46} / \textit{69.87} / \textbf{75.91} / \textbf{86.45} & \textbf{52.60} / \textbf{59.48} / \textbf{67.97} / \textbf{85.00}\\
\bottomrule
\end{tabularx}
\end{table}

\newcommand{\scaleDenoiseSets}{0.14\textwidth}
\renewcommand{\arraystretch}{0.7} %
\begin{figure}[t]
    \centering
    \setlength{\tabcolsep}{1pt}
    \centerline{\begin{tabular}{ccccccc}
     & \footnotesize{\textbf{Ground}} & & & \footnotesize{\textbf{DINOv2 +}} &  & \footnotesize{\textbf{Denoise +}} \\
     
    \footnotesize{\textbf{Input}} & \footnotesize{\textbf{Truth}} & \footnotesize{\textbf{PLAS}} & \footnotesize{\textbf{DINOv2}}  & \footnotesize{\textbf{Registers}} & \footnotesize{\textbf{Denoise}} & \footnotesize{\textbf{Registers}}\\
     
    \includegraphics[width=\scaleDenoiseSets]{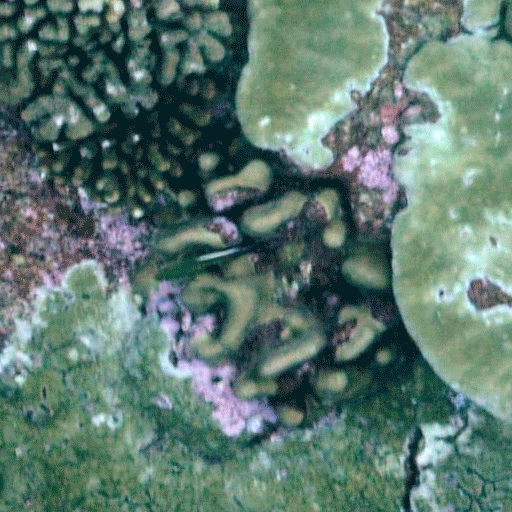} &
    \includegraphics[width=\scaleDenoiseSets]{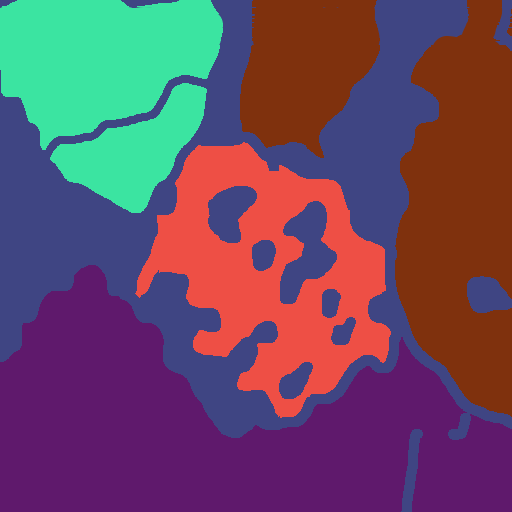} &
    \includegraphics[width=\scaleDenoiseSets]{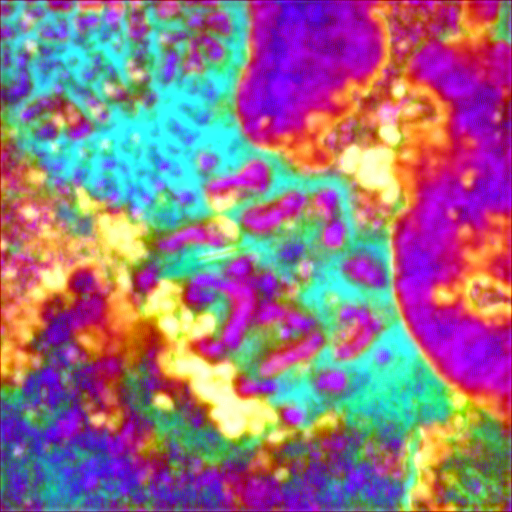} &
    \includegraphics[width=\scaleDenoiseSets]{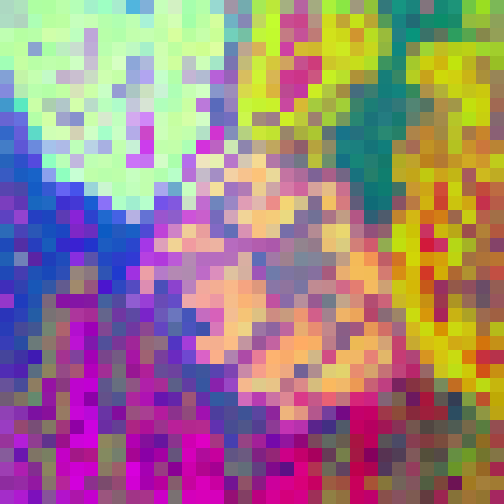} &
    \includegraphics[width=\scaleDenoiseSets]{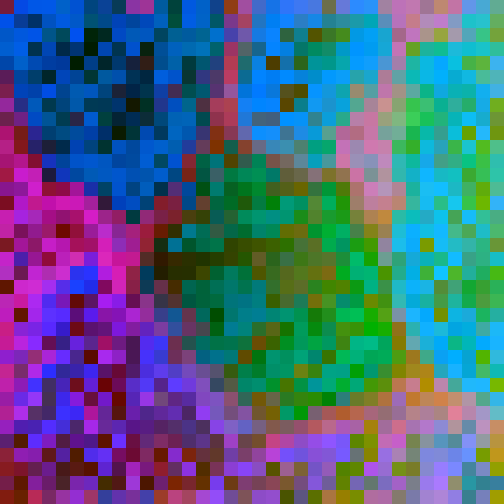} & 
    \includegraphics[width=\scaleDenoiseSets]{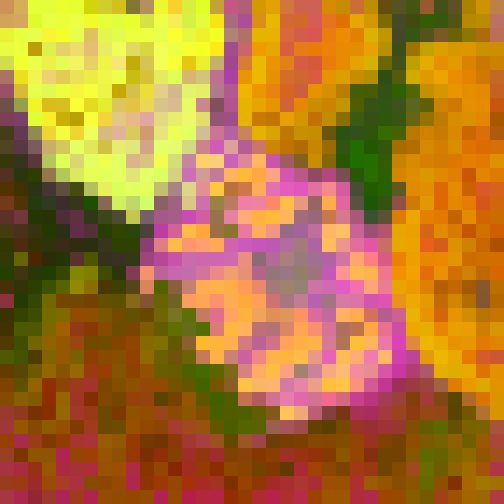}  & 
    \includegraphics[width=\scaleDenoiseSets]{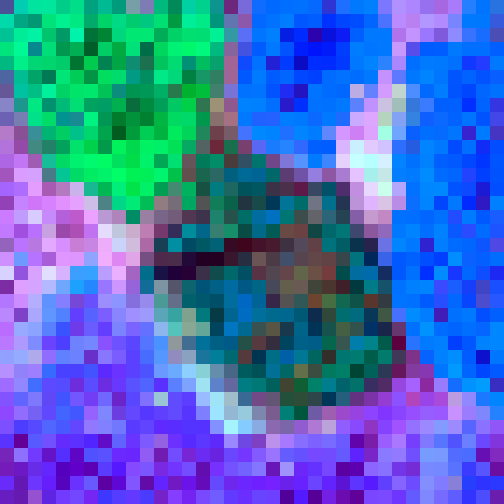} \\    

    \includegraphics[width=\scaleDenoiseSets]{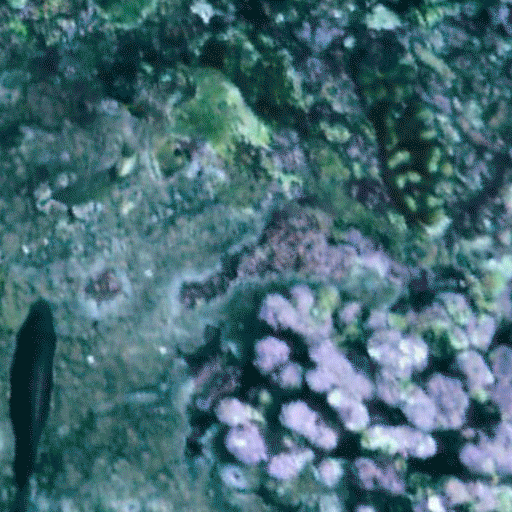} &
    \includegraphics[width=\scaleDenoiseSets]{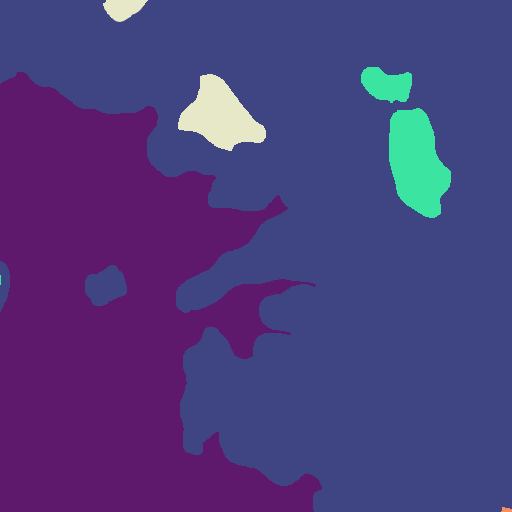} &
    \includegraphics[width=\scaleDenoiseSets]{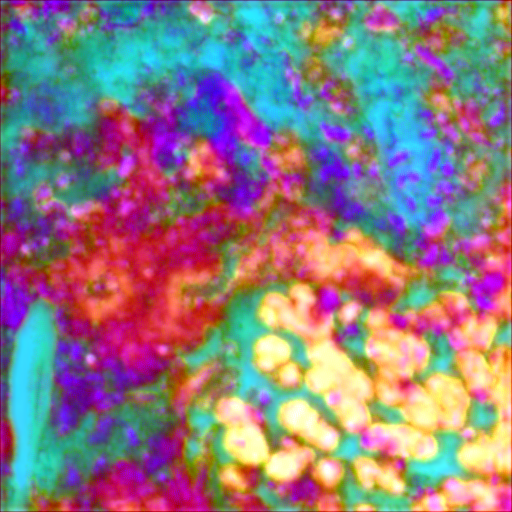} &
    \includegraphics[width=\scaleDenoiseSets]{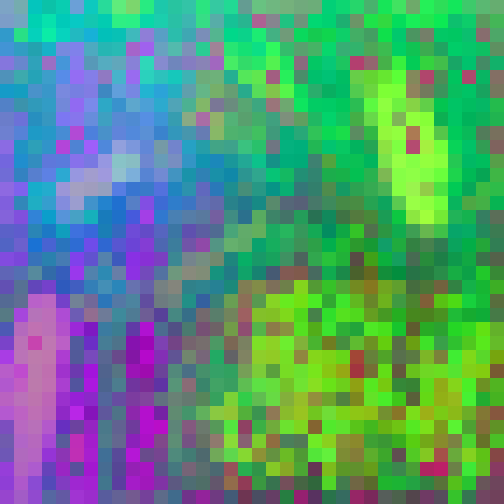} &
    \includegraphics[width=\scaleDenoiseSets]{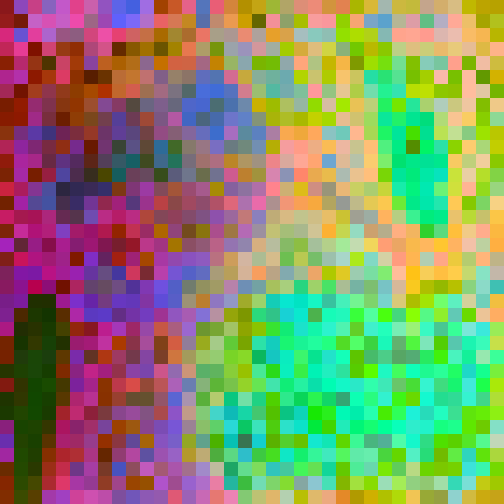} &
    \includegraphics[width=\scaleDenoiseSets]{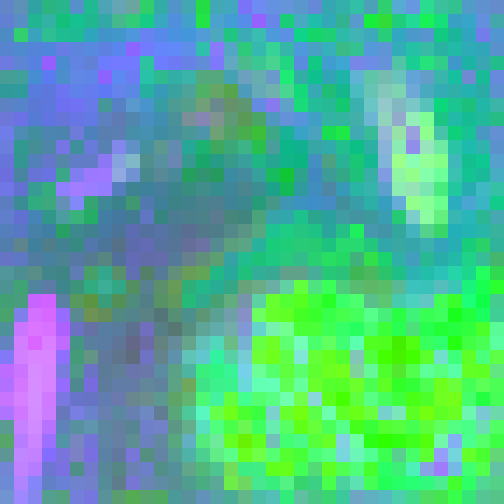} &
    \includegraphics[width=\scaleDenoiseSets]{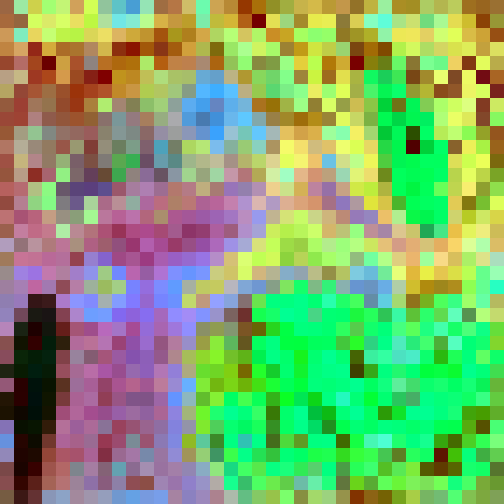} \\    

    \includegraphics[width=\scaleDenoiseSets]{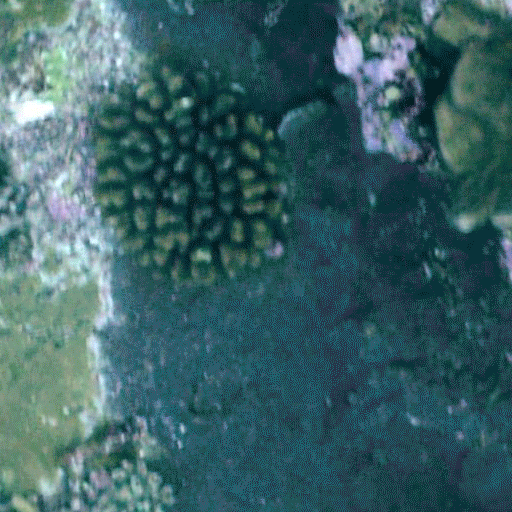} &
    \includegraphics[width=\scaleDenoiseSets]{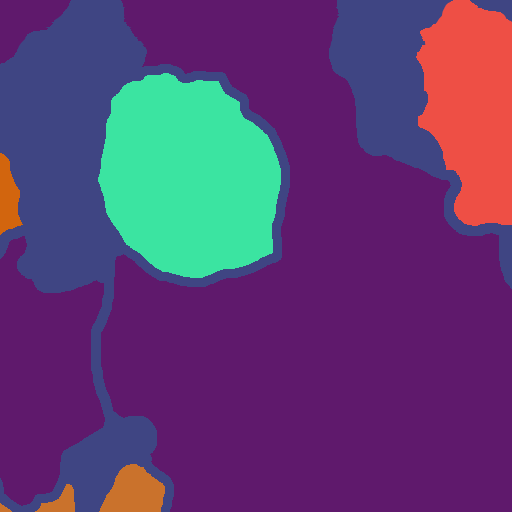} &
    \includegraphics[width=\scaleDenoiseSets]{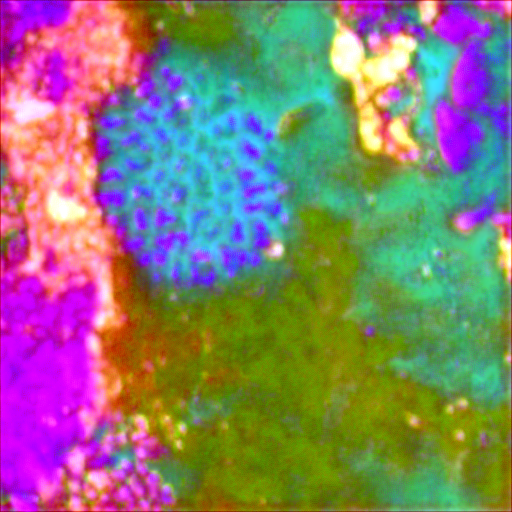} &
    \includegraphics[width=\scaleDenoiseSets]{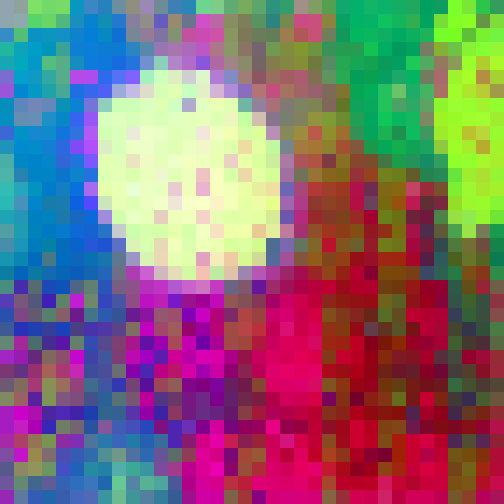} &
    \includegraphics[width=\scaleDenoiseSets]{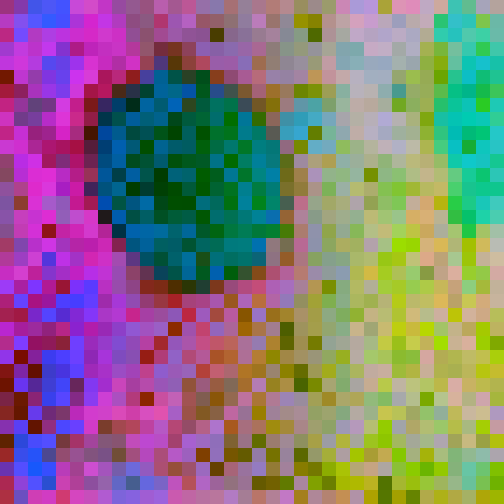} &
    \includegraphics[width=\scaleDenoiseSets]{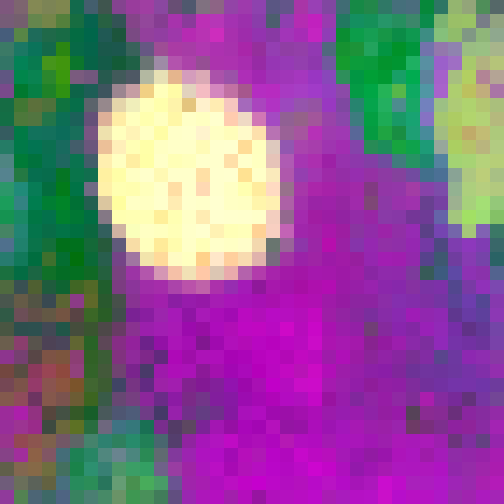} &
    \includegraphics[width=\scaleDenoiseSets]{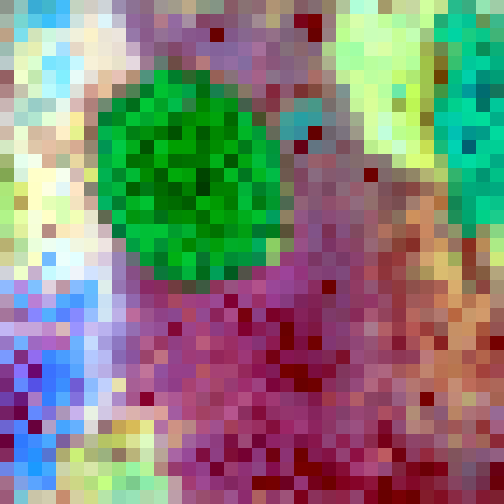} \\   

    \end{tabular}}
    \caption{A comparison is made between Point Label Aware Superpixels (PLAS)~\cite{raine2022point} features, raw DINOv2 features~\cite{oquab2024dinov}, DINOv2 features trained with registers~\cite{darcet2023vision}, denoised DINOv2 features~\cite{yang2024denoising}, and denoised DINOv2 features trained with registers~\cite{yang2024denoising,darcet2023vision} for images in the UCSD Mosaics dataset. In the case of the transformer-based methods, features for each 14x14 pixel patch in the original image are upsampled with bilinear interpolation. Features are visualized in RGB colors by reducing them with Principal Component Analysis (PCA), where the first three components represent the R, G and B channels respectively. Pixels visualized in similar colors indicate similarity in the deep embedding space. The CNN features used by Point Label Aware Superpixels (PLAS)~\cite{raine2022point} fail to cluster pixels into distinct segments which align with the expected ground truth segments. The denoised model significantly reduces artifacts from position embeddings, leading to smoother, cleaner features and better clustering performance.  In the coral imagery context, the models trained with registers do not seem to improve the feature space. Refer to Fig.~\ref{fig:legend} for the color legend.}
    \label{fig:denoise}
\end{figure}

\begin{figure}[t]
\centering
\centerline{\includegraphics[width=150mm, clip, trim=4.5cm 0cm 4.5cm 0cm]{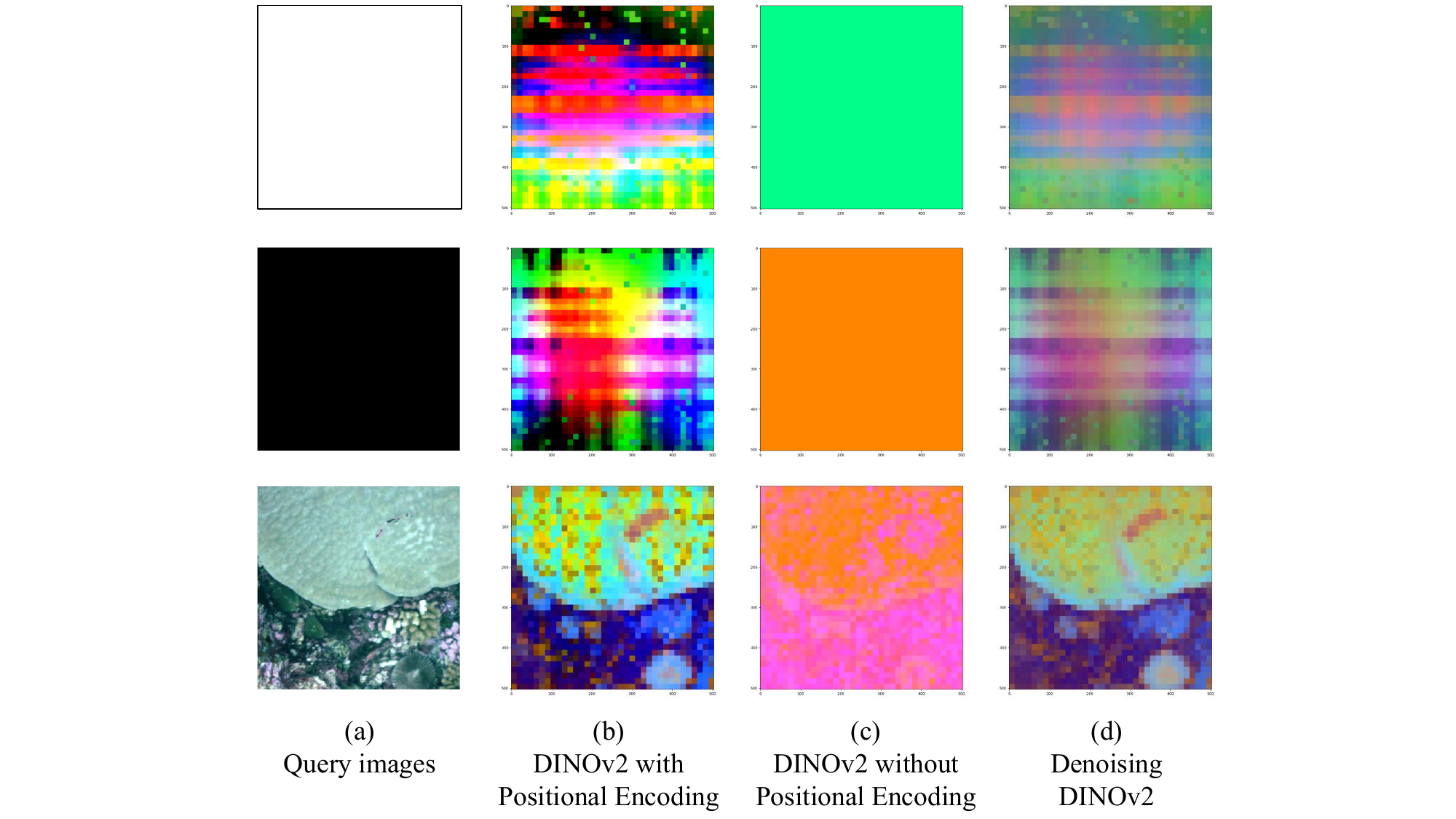}} 
\caption[Noise Artifacts from ViT]{Features extracted from ViTs are subjected to noisy grid-like artifacts caused by the positional encoding in the architecture. Patch tokens are extracted from the DINOv2~\cite{oquab2024dinov} pre-trained ViT, upsampled and visualized into RGB colors using the first three components after reduction with PCA, following the same procedure as in Fig.~\ref{fig:denoise}. Sub-figure a) shows the query images, which are deliberately constant black/white images to demonstrate the artifacts, as well as an example coral image from UCSD Mosaics~\cite{edwards2017large, alonso2019coralseg}, b) shows the features with noise artifacts from the positional encoding, c) shows that the features are free from noise when the positional encoding is not used, and d) shows that the Denoising ViT~\cite{yang2024denoising} significantly reduces the noise artifacts.  Positional encoding is needed during training to inform the spatial relationships between the image patches, enabling the model to effectively learn structure and context in images.}
\label{fig:ViT-noise}
\end{figure}

In Fig.~\ref{fig:denoise}, we present a visualization of the extracted features with the ground truth mask for each image.  We also perform a comparison with a visualization of the CNN features used in the Point Label Aware Superpixel method~\cite{raine2022point}. Further, we demonstrate that the raw DINOv2 features contain artifacts due to the way that DINOv2 is trained with position encoding. These artifacts hinder clustering because there can be multiple individuals of the same species spatially separated within the same image. The impact of the positional embeddings was isolated by visualizing the ViT features for a constant value image, both with and without concatenating the positional embeddings, as seen in Fig.~\ref{fig:ViT-noise}.  The positional embeddings are necessary during training to encode the relative positions of image patches within the original image~\cite{dosovitskiy2020image}.

Training DINOv2 with registers reduces some feature artifacts, but not as effectively as the denoising process. The features obtained through training DINOv2 with registers~\cite{darcet2023vision} and denoising the features~\cite{yang2024denoising} are not as clean as those from the denoised original DINOv2. This is evident in the quantitative results shown in Table~\ref{table:denoise}, where the denoised DINOv2 model achieves the highest performance across the three metrics. The denoised version of DINOv2 minimizes the artifacts, yielding a cleaner feature space and therefore improving the point propagation accuracy (Table~\ref{table:denoise}).  

\begin{figure} 
    \centering
  \subfloat[\mbox{Ablation for $\lambda$}]{%
        \includegraphics[width=0.16\linewidth, clip, trim=0cm 0.7cm 0cm 0.2cm]{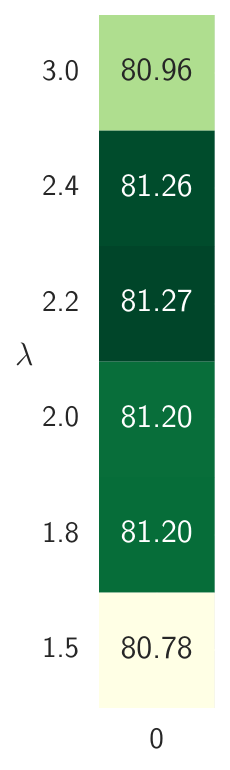}}
        \hspace{2cm}
  \subfloat[Ablation for $\sigma$]{%
        \includegraphics[width=0.16\linewidth, clip, trim=0cm 0.75cm 0cm 0cm]{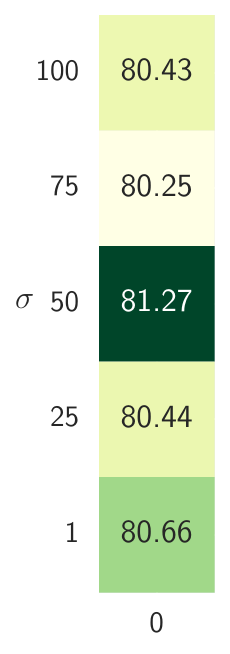}}
  \caption{Label propagation pixel accuracy with 25 point labels, showing that our human-in-the-loop point proposal method is resilient to variation in the value of $\lambda$ and $\sigma$. (a) When the feature similarity map is given more weight, there is a slight improvement in accuracy (we select $\lambda=2.2$). (b) The highest pixel accuracy is achieved if $\sigma=50$ for the Gaussian smoothing of the distance map.}
  \label{fig:ablation} 
\end{figure}

\subsubsection{Weighting the Probability Maps ($\lambda$)}
\label{subsubsec:ablation-weights}
We assess how the $\lambda$ weight, which adjusts the significance of the cosine similarity map (\ref{eq:similarity}) relative to the distance map (\ref{eq:distance}), affects performance. As illustrated in Fig.~\ref{fig:ablation}, a $\lambda$ value of 2.2 yields the highest pixel accuracy. However, our method is relatively insensitive to the precise value of $\lambda$. The analysis indicates that variations in pixel accuracy are less than 1\% across a broad range of $\lambda$ values tested ($1.5 \leq \lambda \leq 3$).

\subsubsection{Exclusion Distance ($\sigma$)}
\label{subsubsec:ablation-distance}
The proposed human-in-the-loop labeling regime accounts for the proximity of labeled pixels by using a Gaussian-smoothed distance mask that measures the distance from all pixels to those that have already been labeled. The Gaussian smoothing introduces the $\sigma$ hyperparameter, which dictates how near new labeled points can be to existing point labels (\ref{eq:gaussian}). Fig.~\ref{fig:ablation} presents an ablation study on the effect of the $\sigma$ hyperparameter on the performance of point label propagation, and we find that our method remains effective across various values of $\sigma$, ranging from 1 to 100.

\subsubsection{Effect of $k$ in KNN}
\label{subsec:ablation-k}

\begin{figure}[t]
    \centering
    \includegraphics[width=0.4\columnwidth]{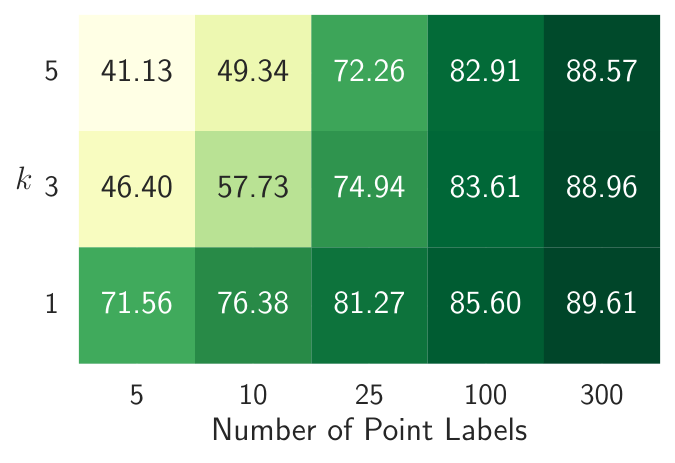}
    \caption{Ablation for $k$. Irrespective of the quantity of points labeled per image, the value $k=1$ always performs the best.}
\label{fig:k-ablation}
\end{figure}

In this ablation study, we thoroughly evaluate different values of $k$ while varying the number of point labels (5, 10, 25, 100, and 300). The results, displayed in Fig.~\ref{fig:k-ablation}, show that the best performance is achieved with $k=1$, which corresponds to using a nearest neighbor classifier. This effect is particularly notable with smaller numbers of point labels, as there are fewer examples available for the clustering algorithm. For instance, with only 5 point labels in an image, it is likely that there is just one labeled point per class, making it ineffective to consider the majority of three or five neighbors, as only one neighbor will correctly represent the class label.

\subsubsection{Effect of Initial Human-Labeled Points}
\label{subsec:ablation-human}

\begin{figure}[t]
\centering
\includegraphics[width=0.4\columnwidth]{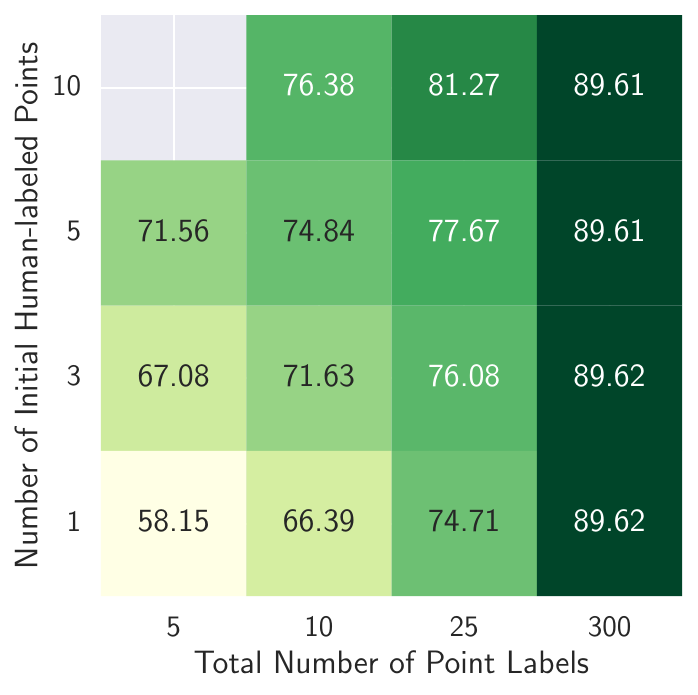}
\caption{Ablation for the number of points labeled initially by the domain expert. Increasing the number of points labeled by the expert results in improved performance; here we choose 10 points, although smaller values still result in comparable performance.}
\label{fig:human-points-ablation}
\end{figure}

Our human-in-the-loop labeling regime starts with 10 points labeled centrally within the largest coral instances. Fig.~\ref{fig:human-points-ablation} explores how varying the number of initial points labeled by a domain expert affects the pixel accuracy of the augmented ground truth masks. We find that increasing the number of initial labels improves the accuracy of point label propagation. However, if minimizing initial labels is desired, similar results can still be achieved with fewer points: using only 3 initial human-labeled points leads to a decrease in pixel accuracy of 4.5\%, 4.8\%, and 5.2\% when there are 5, 10, and 25 total points, respectively.

\subsubsection{Effect of Stage 2 Segmentation Architecture and Backbone}
\label{subsec:ablation-seg}

In this section, we investigate the performance of different semantic segmentation architectures and backbones.  For consistency, we train the models all on our propagated masks, generated using our HIL approach. Table~\ref{tab:seg} compares the performance of the different architectures UNet~\cite{ronneberger2015u} (comprised of 5 blocks), LinkNet~\cite{chaurasia2017linknet} (with ResNet50~\cite{he2016deep} backbone), and Deeplabv3+~\cite{chen2018encoder} on the test dataset. We also compare various backbone architectures for the Deeplabv3+ architecture (ResNet50~\cite{he2016deep}, Inceptionv3~\cite{szegedy2016rethinking} and MobileNetv2~\cite{sandler2018mobilenetv2}). 

\begin{table*}
\caption{Performance of Stage Two: Comparison of Semantic Segmentation Architectures and Backbones (Refer to Section~\ref{subsec:evaluationmetrics} for Metric Definitions).}
\centering
\footnotesize
\centerline{\begin{tabular}{@{}llcccc}

\toprule
\textbf{Architecture} & \textbf{Backbone} & \textbf{PA} & \textbf{mPA} & \textbf{mIoU} & \textbf{Inference Time (fps)} \\

\midrule
UNet~\cite{ronneberger2015u} & n/a & \textbf{80.40} & 49.30 & 40.65 & \textbf{16.74} \\

LinkNet~\cite{chaurasia2017linknet} & ResNet50~\cite{he2016deep} & 68.85 & 44.84 & 36.88 & 12.99 \\

Deeplabv3+~\cite{chen2018encoder} & MobileNetv2~\cite{sandler2018mobilenetv2} & 78.43 & 45.65 & 38.60 & 13.33 \\

Deeplabv3+~\cite{chen2018encoder} & Inceptionv3~\cite{szegedy2016rethinking} & 78.74 & \textbf{51.85} & \textbf{41.28} & 6.70 \\

Deeplabv3+~\cite{chen2018encoder} & ResNet50~\cite{he2016deep} & 74.53 & 41.47 & 32.96 & 13.84 \\
\bottomrule
\end{tabular}}
\label{tab:seg}
\end{table*}

The results in Table~\ref{tab:seg} demonstrate that the Deeplabv3+ with Inceptionv3 backbone outperforms UNet, LinkNet (with ResNet50), and DeepLabv3+ (with ResNet50 and MobileNetv2 backbones) for coral image segmentation.  This is likely due to Inceptionv3's multi-scale convolutional filters within its inception modules, which enable it to capture both local textures and broader contextual information effectively. ResNet50 primarily relies on deep residual connections and sequential convolutions, and MobileNetv2 prioritizes efficiency with depthwise separable convolutions, meaning that Inceptionv3 maintains comparably richer, more diverse feature maps. It is likely that UNet is the next highest performing architecture due to its strong preservation of fine-grained spatial details. UNet’s symmetric encoder-decoder structure, with skip connections at multiple scales, allows it to retain and refine high-resolution features, which is crucial for separating semantically similar coral and marine invertebrate classes.  While LinkNet also incorporates skip connections, it has a lightweight decoder which is less capable of restoring intricate boundaries compared to UNet’s more expansive upsampling path.

For consistency with prior approaches, the Deeplabv3+ architecture with ResNet50 backbone is used to evaluate the impact of different point label propagation approaches .

Despite the Deeplabv3+ with Inceptionv3 backbone achieving higher performance on the semantic segmentation task, the Deeplabv3+ architecture with ResNet50 backbone is used to evaluate the impact of different point label propagation approaches (Table~\ref{tab:stage2}), to maintain consistency with prior approaches~\cite{pierce2020reducing, raine2022point}.

\subsection{Effect of Point Label Quantity}
\label{subsec:results-label-quantity}

Increasing the number of point labels led to better performance in the point label propagation task (Fig.~\ref{fig:point-graph}). Having access to enough point labels is particularly important for the Fast MSS~\cite{pierce2020reducing} method. If an image has been annotated with grid-spaced points, Fast MSS shows a significant improvement in mIoU, rising from 7.5\% to 86.4\% as the number of labels increases from 5 to 300, a difference of 78.9\%. In comparison, Point Label Aware Superpixels~\cite{raine2022point} and our DINOv2 and KNN methods show improvements of 56.5\% and 47.1\%, respectively (Table~\ref{tab:stage1}).

While all methods benefit from an increase in point labels, the rate of improvement diminishes as the number of labels grows from 100 to 300. For instance, when increasing grid labels from 100 to 300 points per image, the Fast MSS~\cite{pierce2020reducing} approach improves by 11.7\% in mIoU, compared to a 67.3\% improvement when increasing from 5 to 100 points. Similarly, the Point Label Aware Superpixels show a 56.4\% improvement in mIoU when moving from 5 to 100 grid points and a 9.5\% improvement from 100 to 300 points. In the case of the denoised DINOv2 and KNN, the mIoU improves by 26.2\% from 5 to 100 HIL points and by 6.2\% from 100 to 300 HIL points.

\subsection{Effect of Point Label Placement Style}
\label{subsec:results-label-placement}

All the methods evaluated show advantages when using grid placement of point labels compared to random placement (Fig.~\ref{fig:point-graph}). This effect is especially notable with the multi-level superpixels (Fast MSS)~\cite{pierce2020reducing}, which shows significant absolute improvements in mIoU with grid-spaced labels over random labels: 13.3\%, 9.9\%, 6.8\% and 6.3\% for 50, 100, 200 and 300 points, respectively. The Point Label Superpixels also benefit from grid spacing, with improvements of 8.4\%, 4.8\%, 4.4\% and 3.9\% for the same label quantities. Grid-spaced labels ensure uniform coverage across the entire image and make optimal use of each label. As illustrated in Fig.~\ref{fig:qualitativeresults}, randomly placed labels can be clustered closely together, diminishing the amount of useful information.

Fig.~\ref{fig:qualitativeresults} also shows that, with very few point labels (5 to 10 per image), there is considerable benefit from utilizing domain expert knowledge to select points centrally within instances. Further pixels can then be iteratively selected using the point propagation model described in Section~\ref{sec:method}. The augmented ground truth masks produced by our approach (top two rows of Fig.~\ref{fig:qualitativeresults}) are significantly closer to the actual ground truth compared to previous methods. We plan to explore applying our human-in-the-loop labeling regime to other techniques in future research.

When labels are sparse, multi-level superpixel methods~\cite{pierce2020reducing, alonso2019coralseg} struggle because they depend on layering labeled regions from various scales. Similarly, the point label superpixel method faces challenges in sparse cases, as its superpixel boundaries are not forced to align with instance boundaries by the conflicting point labels~\cite{raine2022point}. Our method performs well with sparse labels because it assigns the correct class to pixels even if they are spatially distant from labeled points, by relying on clustering in the deep feature space.

\section{Discussion}
\label{sec:discussion}

\subsection{Impact}
\label{subsec:impact}

The core impact of this work is in reducing the cost associated with expert annotation of coral reef imagery. Labeling images for semantic segmentation is expensive, as domain experts must manually annotate each point, with costs accumulating per labeled point. By significantly decreasing the number of required point labels while maintaining or improving segmentation accuracy, our approach leads to direct cost savings. For example, if traditional methods require 300 point labels per image and our method achieves comparable performance with only 5 point labels, this represents a 98.3\% reduction in labeling effort and therefore cost. Typically, datasets should be in the thousands of images to enable sufficient generalization to a range of visual conditions -- this translates into significant financial savings, making model fine-tuning and adaptation for large-scale coral reef monitoring more cost-effective and feasible.

This paper presents a comprehensive set of experiments analyzing how the quantity of point labels influences segmentation performance. Given that the definition of satisfactory segmentation is subjective and varies based on user needs, we evaluate multiple labeling schemes across different levels of human input. Our results illustrate the trade-offs between annotation effort and segmentation accuracy, showing that while additional point labels generally improve performance, highly accurate point label propagation with fewer point labels can still be obtained by using our proposed approach (Fig.~\ref{fig:point-graph}). The number of point labels should be selected based on the desired use case, the dataset characteristics, labeling time and budget, and the availability of expert annotators. Rather than prescribing a fixed number of necessary labels, we provide insights into how different annotation strategies impact outcomes, allowing users to make informed decisions based on their specific constraints and application requirements.

\subsection{Limitations}
\label{subsec:limitations}

Our approach is limited by the two stages in the architecture \ie one stage of processing to generate the augmented ground truth masks for the training data, followed by another stage for training the segmentation model on the augmented masks.  It would be more efficient to employ a single stage approach which takes as input the underwater image and sparse point labels, and outputs dense segmentation masks at inference time. A single stage approach was trialled in which deep features for labelled pixels were stored in a memory bank and used as templates for clustering pixels in new imagery. However, the vision transformer architecture encodes global information in each embedding, meaning that pixels of the same class in different images are far apart in the deep embedding space. A single-stage approach which overcomes this limitation is left for future work. 

The human-in-the-loop labeling regime in this work was simulated using the ground truth masks with the UCSD Mosaics dataset~\cite{edwards2017large, alonso2019coralseg}.  The regime starts by the `domain expert' labelling up to ten pixels centrally in coral instances, and then iterative prediction of the optimal locations for labelling the next points up to a pre-defined maximum quantity of point labels. This was simulated by obtaining the ground truth label at the identified pixel location as though it was labelled by an expert `in-the-loop'.  The approach would benefit from user testing with multiple domain experts to determine how inter-observer variability impacts the performance of the approach.

\subsection{Future Work}
\label{subsec:futurework}

Our method could be extended to treat the human-in-the-loop scenario as an active learning problem during deployment. While our current human-in-the-loop framework is focused on point label propagation, it could be modified to detect novel species in real-time and utilize domain expert input to incorporate new classes through active learning. In this scenario, a deep learning model could be deployed on a robotic platform and used to predict previously unseen targets in the footage.  The human-in-the-loop scheme would then present identified anomalies to a domain expert in real-time. The expert determines whether the class is of interest and if so, specifies the class label. As further examples are detected, the examples could be incorporated in an active learning framework where the model iteratively incorporates additional classes.  Underwater anomaly detection could have significant implications for species discovery and would enable models to be quickly adapted to new locations and species encountered, rather than exhibiting degraded performance when deployed in a different setting from the training data. 

Another future direction could be to obtain features usable across different images. The architecture and training of the DINOv2 foundation model incorporates global image information into per-pixel deep features, resulting in deep features of the same coral species from different images not being similar in the deep embedding space. Enabling feature similarity across different images could eliminate the need for the second stage of this architecture, thereby simplifying the framework by removing the need for DeepLabv3+ training.

The human-in-the-loop labeling regime was simulated using the ground truth masks with the UCSD Mosaics dataset~\cite{edwards2017large, alonso2019coralseg}. Future research might include testing with multiple domain experts to assess how experts interact with the human-in-the-loop regime and how the inter-observer variability affects performance.

Given the increasing frequency of global coral bleaching events, there is a growing need for detailed monitoring of coral reefs and the identification of temperature-resistant species. Future research could expand our approach to enable precise recognition of coral reef health indicators. 

This work has focused on segmentation of coral imagery, however the approach presented could be used for broad-scale surveys of other species of interest, including seagrass meadows and algae.  Further, the Stony Coral Tissue Loss Disease is an aggressive disease which can kill entire colonies of stony corals in a period of months~\cite{papke2024stony}.  The spread of the disease is not fully understood so underwater monitoring over varying spatial and temporal scales is important for detection and management of the disease~\cite{combs2021quantifying, papke2024stony}.  As there is limited data available which captures the visual characteristics of the disease, the human-in-the-loop labeling regime presented in this work could be used to enable fast annotation, training and deployment of disease detection models. 

In addition to marine surveys, environmental monitoring on land is often subject to similar data availability and annotation challenges. Broad-scale surveys completed using Unmanned Aerial Vehicles (UAVs) or satellite imagery require annotation for training deep learning models, and in domain-specific applications this can be expensive and time-consuming, as it is for underwater imagery.  Future work could investigate the application of our presented human-in-the-loop labeling approach on aerial and remote data for domain-specific environmental monitoring tasks.

\section{Conclusion}
\label{sec:conclusions}
Marine ecosystem monitoring requires efficient analysis of vast coral reef imagery, but expert annotation for semantic segmentation is prohibitively expensive and time-consuming, creating a critical bottleneck for large-scale ecological studies.
This work has demonstrated for the first time how general foundation models can be effectively utilized for point label propagation in domain-specific underwater imagery without requiring fine-tuning, combined with a novel strategic human-in-the-loop labeling framework that selects annotation locations based on feature uncertainty and spatial diversity.

By leveraging denoised DINOv2 features with a straightforward KNN algorithm, we generate augmented ground truth masks from extremely sparse labels (5-25 points per image) through a two-stage pipeline that first propagates point labels and then trains semantic segmentation models on the resulting augmented masks. When combined with our human-in-the-loop labeling approach, we achieve significant improvements in mIoU for the label propagation task: 19.7\%, 18.3\%, and 8.2\% for 5, 10, and 25 point labels, respectively, when compared to previous state-of-the-art methods. These performance gains are evident in the semantic segmentation task as well: Training a DeepLabv3+ model on augmented ground truth masks created with DINOv2, KNN, and our human-in-the-loop labeling approach yields improvements of 8.8\% in pixel accuracy and 13.5\% in mIoU with just 5 point labels.

The large number of coral species (there are an estimated 800 species of hard corals alone~\cite{dietzel2021population}), visual complexity of coral imagery, and geographic diversity of species mean that real-world deployment of underwater computer vision must involve data collection in the target environment, efficient annotation, and then model training or fine-tuning. The method proposed in this work contributes to the goal of rapid training and deployment of location-specific models by significantly reducing the amount of required annotations while achieving accurate point label propagation, with broader implications for cost-effective computer vision in specialized domains where expert annotation is scarce and expensive.

\section*{\normalsize Acknowledgments}
This work was done in collaboration between QUT and CSIRO Data61. S.R., F.M., N.S., and T.F.~acknowledge continued support from the Queensland University of Technology (QUT) through the Centre for Robotics. T.F.~acknowledges funding from an Australian Research Council (ARC) Discovery Early Career Researcher Award (DECRA) Fellowship DE240100149. S.R.~and T.F.~acknowledge support from the Reef Restoration and Adaptation Program (RRAP) which is funded by a partnership between the Australian Government’s Reef Trust and the Great Barrier Reef Foundation.  Computational resources and services used in this work were provided by the eResearch Office, Queensland University of Technology, Brisbane, Australia.
\bibliographystyle{IEEEtran}
\bibliography{main}

\end{document}